\documentclass[preprint,12pt,nopreprintline]{elsarticle}

\usepackage{amsmath,amssymb}
\usepackage{graphicx}
\usepackage{multirow}
\usepackage{booktabs}
\usepackage{array}
\usepackage{hyperref}
\usepackage{subcaption}
\usepackage{tikz}
\usepackage{float}
\usepackage{placeins}
\usetikzlibrary{arrows.meta, positioning, calc}
\newcommand{\qualcell}[3]{%
  \begin{minipage}[t]{0.235\textwidth}
    \centering
    {\small #3\par}
    \vspace{3pt}
    \begin{minipage}[c][0.29\textheight][c]{\linewidth}
      \centering
      \includegraphics[width=\linewidth,height=0.29\textheight,keepaspectratio]{#1}
    \end{minipage}\par
    \vspace{3pt}
    {\small #2\par}
  \end{minipage}%
}

\journal{ISPRS Journal of Photogrammetry and Remote Sensing}

\makeatletter
\gdef\emailauthor#1#2{\stepcounter{ead}%
     \g@addto@macro\@elseads{\raggedright%
      \let\corref\@gobble\def\@@tmp{#1}%
      \eadsep{\ttfamily\expandafter\strip@prefix\meaning\@@tmp}%
      \def\eadsep{\unskip,\space}}%
}
\makeatother

\begin{document}

\begin{frontmatter}

\title{MONO‑HYDRA++: Real‑Time Monocular Scene Graph Construction with
Multi‑Task Learning for 3D Indoor Mapping}

\author[inst1]{\texorpdfstring{U.V.B.L. Udugama\corref{cor1}}{U.V.B.L. Udugama}}
\ead{b.udugama@utwente.nl}
\author[inst1]{George Vosselman}
\ead{george.vosselman@utwente.nl}
\author[inst1]{Francesco Nex}
\ead{f.nex@utwente.nl}

\affiliation[inst1]{organization={Department of Earth Observation Science, University of Twente},%
city={Enschede}, postcode={7522 NH}, country={The Netherlands}}

\cortext[cor1]{Corresponding author}

\begin{abstract}
Autonomous agile robots need more than metric geometry: they must understand objects,
rooms, places, and spatial relations to support intelligent behavior in tasks such as
search, inspection, exploration, and human-robot interaction. Conventional metric maps
enable localization and collision-free navigation, but they do not provide this
semantic and relational structure. 3D scene graphs fill this gap by connecting geometry
with object-level and room-level understanding.

Building such representations on agile platforms requires balancing sensing, payload,
power, and computation. We present Mono-Hydra++, a real-time monocular RGB+IMU pipeline for indoor
metric-semantic mapping and hierarchical 3D scene graph construction. The system combines
M2H-MX, our multi-task depth and semantic predictor, with RVIO2
visual-inertial odometry (VIO) using deep-feature tracking, local graph refinement using sparse predicted-depth
constraints and semantic filtering, and pose-aware temporal alignment before volumetric
fusion in the Mono-Hydra backend. On the Go-SLAM ScanNet evaluation subset,
Mono-Hydra++ achieves 1.6\% lower average trajectory error (ATE) than the strongest
RGB-D baseline in our comparison, while using only monocular RGB+IMU input. On calibrated
7-Scenes, it improves average ATE by 29.8\% over the strongest competing calibrated
baseline. We further evaluate Mono-Hydra++ in a real ITC building deployment using
RealSense RGB+IMU and demonstrate embedded feasibility by deploying the ONNX/TensorRT
FP16 M2H-MX-L perception model at about 25 FPS with $256\times192$ input on a Jetson Orin NX 16GB.
\end{abstract}


\end{frontmatter}

\section{Introduction}

Understanding indoor 3D environments is a core challenge in photogrammetry, computer
vision, and robotics. Traditional SLAM systems primarily reconstruct scene geometry,
typically in the form of point clouds or meshes. While these representations are useful
for localization, visualization, and geometric mapping, they are not directly
actionable for higher-level robotic reasoning because they do not explicitly capture
objects, rooms, and their spatial relationships. In real-world scenarios such as
retrieving an object from a specific room, searching for a target item in a warehouse,
or locating victims in a collapsed building, robots must interpret the environment in
a more structured and task-oriented manner. 3D scene graphs address this need by
organizing metric, semantic, and topological information in a hierarchical
representation~\cite{armeni20193d}, enabling reasoning beyond raw geometry, and have
shown practical value in semantic mapping, task planning, and human--robot
interaction~\cite{rosinol2020kimera,hughes2022hydra}.

This form of structured spatial understanding is especially important for agile
robotic platforms, where perception must support both navigation and decision making
under strict payload, power, and computational constraints. Real-time scene graph
pipelines such as Kimera~\cite{rosinol2021kimera} and Hydra~\cite{hughes2022hydra}
provide a foundation for structured spatial reconstruction. Here, we investigate
real-time construction of metric-semantic 3D scene graphs from monocular RGB and an
inertial measurement unit (IMU), without a dedicated depth sensor.

Monocular cameras paired with IMUs are attractive for this setting, yet monocular scene
graph construction remains difficult: depth estimation is ill-posed in low-texture
indoor scenes, visual-inertial odometry (VIO) is fragile when features are sparse, and
frame-wise predictions introduce temporal flicker that degrades mapping and graph
consistency~\cite{godard2019digging,huai2022robocentric,liu2022efficient}. These errors
affect both pose estimation and the geometry and labels integrated into the map.

Multi-task learning (MTL) improves dense scene understanding by sharing structure
across depth, semantics, normals, and edges. Methods such as PAD-Net~\cite{xu2018pad},
MTI-Net~\cite{vandenhende2020mti}, InvPT~\cite{ye2022inverted}, and
MTFormer~\cite{xu2022mtformer} benefit from sharing features across tasks. For mapping, however,
agreement between tasks within an image does not by itself ensure agreement between
views of the same surface. Frame-wise depth errors can place that surface at different
3D locations, while changing semantic labels can assign conflicting classes to it.
Fusion then receives inconsistent geometric and semantic evidence, even when individual
predictions are plausible. Temporal alignment combines predictions across views before
map fusion.

Mono-Hydra~\cite{udugama2023monohydra}, our earlier work, combined a single-task
semantic network with a monocular depth estimator. However, this two-network design
led to inconsistencies between geometry and semantics.

Building on this, we introduced M2H~\cite{udugama2025m2h}, a unified multi-task
learning framework that jointly predicts depth, semantics, surface normals, and edges.
This improved mesh completeness, semantic boundary sharpness, and scene graph
accuracy, while also reducing computational overhead.

In this paper, we present Mono-Hydra++, a real-time monocular RGB+IMU pipeline that
uses M2H-MX~\cite{udugama2026m2hmx}, our previously introduced
multi-task dense predictor, to supply depth maps and semantic labels. The pipeline
couples these outputs with visual-inertial odometry, pose-aware temporal fusion, and
hierarchical 3D scene graph construction.

Mono-Hydra++ addresses three gaps that remain when learned monocular perception is used
to build 3D scene graphs:
\begin{itemize}
    \item \textbf{Lightweight monocular metric-semantic mapping:} Mono-Hydra++
          reconstructs metric-semantic meshes and hierarchical 3D scene graphs from
          monocular RGB+IMU input, without a dedicated depth sensor. These
          structured representations support indoor mapping and robotic reasoning,
          and the perception model runs in real time on an embedded GPU.

    \item \textbf{Geometry-aware perception-to-VIO coupling:} Mono-Hydra++ introduces
          local graph refinement that integrates M2H-MX predictions into the
          visual-inertial mapping pipeline rather than using them only as frame-wise
          dense outputs. Sparse predicted-depth factors constrain tracked landmark
          depths, while semantic masks exclude observations on dynamic classes from
          the local graph. To support feature association during rapid rotation and in
          weakly textured indoor scenes, we augment RVIO2's tracker with SuperPoint
          descriptor matching around IMU-predicted locations, followed by Lucas--Kanade
          alignment.

    \item \textbf{Pose-aware temporal fusion before mapping:} To limit
          depth and label disagreements across views, Mono-Hydra++ aligns recent depth
          and semantic predictions using VIO poses before
          volumetric fusion into a shared 3D voxel map. A moderate temporal window
          gives the lowest trajectory error, and a short window gives the best
          reconstruction and scene-graph accuracy on the longer Office sequences.
\end{itemize}

\begin{figure}[!t]
\centering
\includegraphics[width=0.98\textwidth]{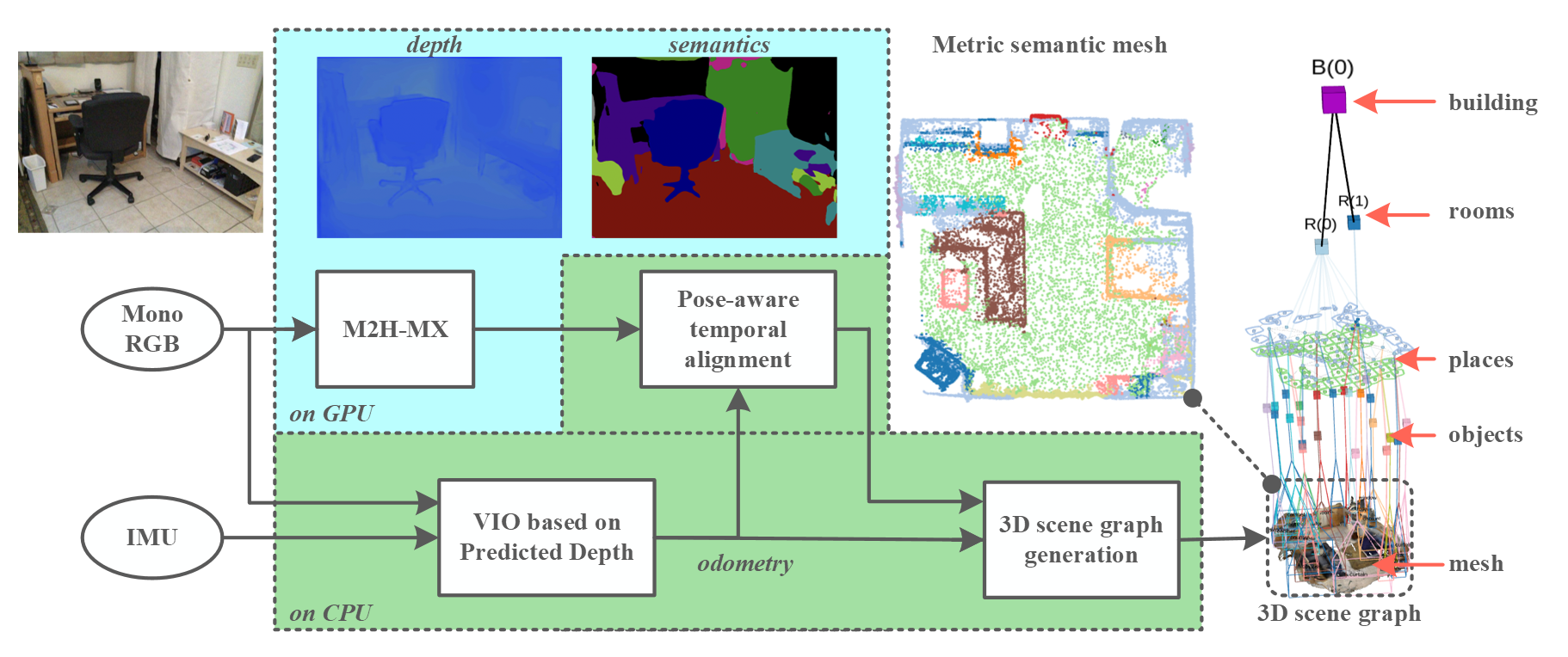}
\caption{System outputs and data flow of Mono-Hydra++. M2H-MX predicts depth and
semantics from monocular RGB on the GPU, and visual-inertial odometry uses the predicted
depth together with IMU measurements. Pose-aware temporal alignment uses the odometry to
stabilize the predictions before 3D scene graph generation, which produces a
metric-semantic mesh and a hierarchical 3D scene graph with building, room, place, and
object nodes. The detailed architecture is shown in Fig.~\ref{fig:pipeline_overview}.}
\label{fig:scenegraph_outputs}
\end{figure}

Fig.~\ref{fig:scenegraph_outputs} summarizes how these components produce the final metric-semantic mesh and hierarchical 3D scene graph.

\section{Related Work}
\label{sec:related}

Prior work relevant to our approach spans (i) multi-task and dense prediction,
(ii) monocular mapping and depth/VIO, and (iii) 3D scene graphs for robotics.

\subsection{Multi-Task and Dense Prediction}
Multi-task models share features and exchange task-specific information to jointly
predict depth and semantics, as in PAD-Net~\cite{xu2018pad},
MTI-Net~\cite{vandenhende2020mti}, and InvPT~\cite{ye2022inverted}. Our earlier
M2H~\cite{udugama2025m2h} and M2H-MX~\cite{udugama2026m2hmx}
provide dense perception for monocular mapping. We reuse M2H-MX as the perception module.

\subsection{VIO and Mapping}
Visual--inertial odometry (VIO) estimates camera motion, while dense mapping determines
how scene geometry is reconstructed and updated over time. Traditional monocular SLAM
pipelines rely on feature-based or direct tracking with IMU pre-integration, whereas
learning-based methods increasingly incorporate deep features or learned scene
representations.

\paragraph{Visual--Inertial Odometry (VIO)}
Combining IMU and monocular vision improves robustness, particularly in low-texture
indoor environments. Classical VIO pipelines such as ORB-SLAM3~\cite{campos2021orb}
and VINS-Mono~\cite{qin2018vins} fuse visual keypoints and IMU measurements in tightly
coupled optimization frameworks. Square-root robocentric VIO methods such as
R-VIO2~\cite{huai2022square} instead maintain the estimator in a robocentric frame and
apply QR-based square-root updates, improving numerical stability and efficiency for
monocular camera--IMU odometry. Recent deep VIO methods~\cite{han2019deepvio} learn to
fuse visual descriptors and inertial features, sometimes through selective gating or
transformer-based temporal fusion~\cite{fei2024transformer}. Online adaptive VIO
methods~\cite{pan2024adaptive} further mitigate domain shift. Recent hybrid SLAM systems
extend this direction by incorporating learned geometric priors into optimization.
VGGT-SLAM~\cite{maggio2025vggt-slam} leverages feed-forward transformer-based geometric
features to improve correspondence robustness, while VGGT-SLAM 2.0 further improves
submap alignment, loop closure verification, and trajectory accuracy through a revised
factor-graph formulation~\cite{maggio2026vggtslam20realtimedense}. MASt3R-SLAM~\cite{murai2024_mast3rslam}
similarly uses dense matching priors to strengthen geometric constraints in low-texture
regions. These systems show that learned geometric priors can reduce drift and improve
trajectory stability in challenging monocular settings. Mono-Hydra~\cite{udugama2023monohydra}
integrates a monocular depth estimator with robocentric VIO, achieving metric scale and
stable real-time performance.

\paragraph{Dense Mapping and Neural SLAM}
Neural implicit mapping has enabled dense, high-fidelity reconstruction beyond
traditional truncated signed distance function (TSDF)- or surfel-based systems. iMAP~\cite{Sucar:etal:ICCV2021} showed that a single MLP
can serve as the only scene representation in a real-time RGB-D SLAM system, but its
limited network capacity restricts scalability. NeRF-SLAM~\cite{rosinol2023nerf} brought
this direction to monocular input by combining dense monocular tracking with a hash-grid
radiance field, and NICER-SLAM~\cite{zhu2024nicer} jointly optimizes camera poses and a
hierarchical neural implicit map from RGB input using monocular geometric cues, while Vox-Fusion~\cite{yang2022vox} and
ESLAM~\cite{johari2023eslam} further improved efficiency and geometric fidelity.
SP-SLAM~\cite{hong2025sp} later analyzed limitations such as fixed-resolution signed
distance function (SDF) grids and keyframe bundles, and proposed more scalable
implicit structures. Despite this progress, neural implicit SLAM methods remain computationally
heavy and GPU intensive for real-time deployment.

More recently, Gaussian Splatting-based SLAM methods~\cite{yugay2024gaussianslamphotorealisticdenseslam,sandstrom2024splat}
have emerged as an efficient alternative to NeRF-style representations, modeling scenes
as collections of 3D Gaussians optimized via differentiable rendering. They offer
high-fidelity dense mapping with much faster rendering, and recent variants support
online SLAM for real-time or near real-time operation. However, these methods remain
focused primarily on photometric and geometric reconstruction, typically require
substantial GPU resources, and do not explicitly model semantic structure such as
objects, rooms, and spatial relationships.
For this paper, the more direct comparison is with real-time dense SLAM systems that
already couple tracking and mapping.

RGB-D dense SLAM systems such as iMAP~\cite{Sucar:etal:ICCV2021},
NICE-SLAM~\cite{Zhu2022CVPR}, DROID-SLAM~\cite{teed2021droid}, and
Go-SLAM~\cite{zhang2023goslam} are strong baselines for trajectory accuracy and geometry,
but they assume depth input. Monocular variants (e.g., DROID-SLAM in monocular
mode) can suffer scale drift and reduced reconstruction stability. Mono-Hydra++ targets
this gap by coupling multi-task monocular predictions with VIO and temporal alignment to
approach RGB-D performance using only RGB+IMU and produce structured
semantic representations for downstream scene-graph reasoning.

\subsection{3D Scene Graphs and Semantic Mapping}
3D scene graphs connect metric geometry to objects, rooms, and spatial relations.
Armeni et al.~\cite{armeni20193d} organize semantic and spatial information in a
hierarchy that also relates camera observations to the scene, so that an object can be
associated with a room and interpreted through its relations to other scene elements.
SceneGraphFusion~\cite{wu2021scenegraphfusion} predicts such graphs incrementally from
RGB-D sequences, using a graph neural network to infer object semantics and relations
from geometric segments.

Kimera~\cite{rosinol2020kimera} combines metric-semantic reconstruction with
visual-inertial estimation, and its Dynamic Scene Graph extension~\cite{rosinol2021kimera}
builds a layered representation of static and dynamic entities from visual-inertial
data. Hydra~\cite{hughes2022hydra} constructs and optimizes the scene graph incrementally
during exploration: it extracts places from a local Euclidean signed distance field,
groups them into rooms, and propagates loop-closure corrections across the layers
through deformation-based optimization.

Other work couples scene structure more tightly to estimation or broadens the semantics.
S-Graphs+~\cite{bavle2023sgraphsplus} jointly optimizes keyframes, wall planes, rooms, and
floors in a hierarchical factor graph from 3D LiDAR, and
vS-Graphs~\cite{tourani2026vsgraphs} brings this coupling to RGB-D visual SLAM.
Open-vocabulary methods such as ConceptGraphs~\cite{gu2024conceptgraphs} and
HOV-SG~\cite{werby2024hovsg} fuse foundation-model outputs into object-centric or
floor--room--object graphs for language-guided reasoning and navigation, while
Clio~\cite{maggio2024clio} groups object and place primitives into task-relevant
objects and regions from natural-language task descriptions. Most of these systems use
RGB-D or LiDAR input.

Mono-Hydra~\cite{udugama2023monohydra} supplies learned depth and semantics to a
monocular scene-graph pipeline. We retain its hierarchical backend and address errors
in the predicted depth, semantic labels, and poses used to construct the graph. Sparse
depth factors and semantic masking refine the local graph, while temporal alignment
brings predictions from recent views into the current camera frame before map fusion.
We evaluate the resulting trajectories, meshes, and object nodes.

\section{Methodology}
\label{sec:method}

\begin{figure*}[!t]
  \centering
  \includegraphics[width=\textwidth]{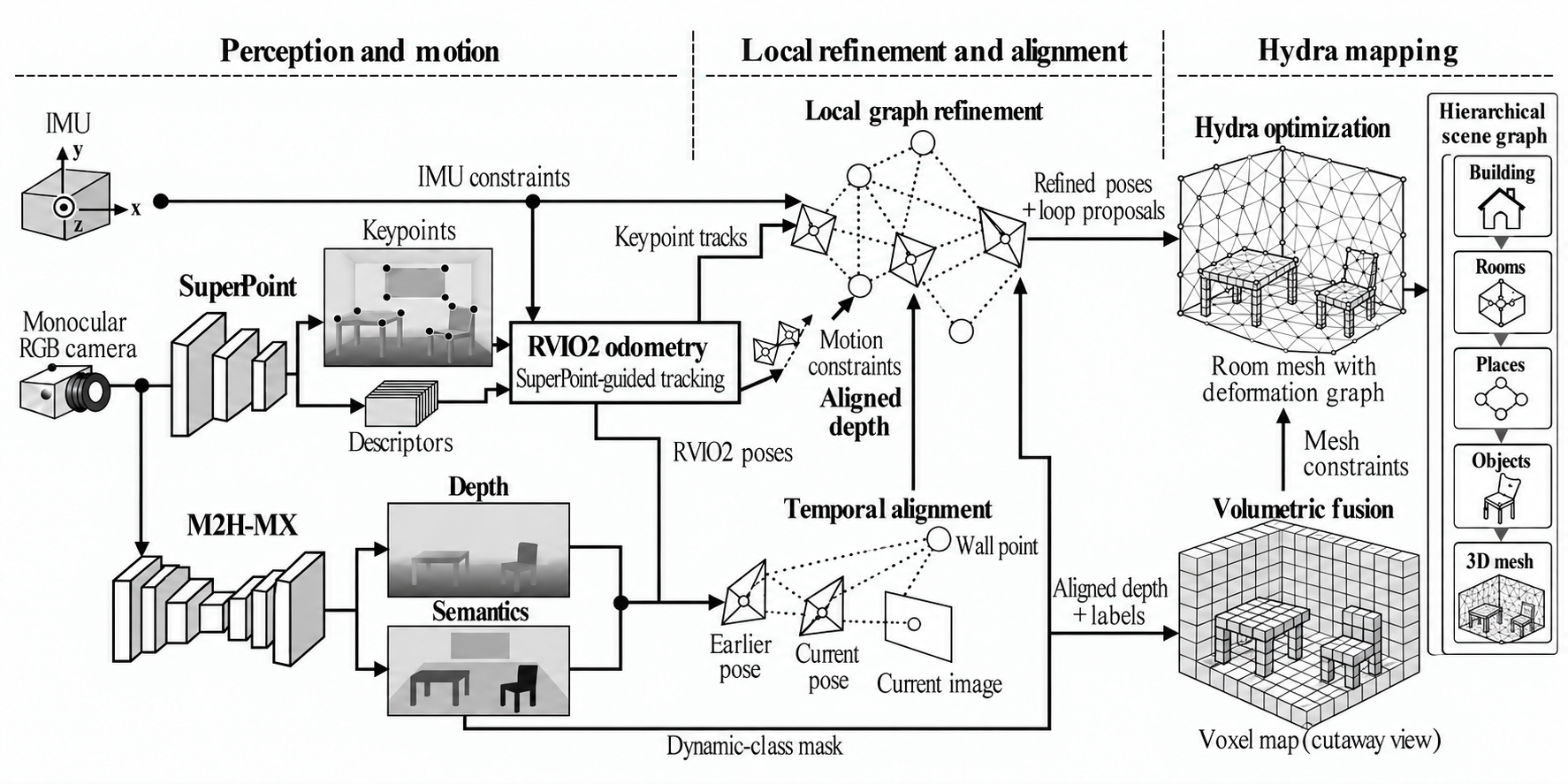}
  \caption{System overview of Mono-Hydra++. A monocular RGB camera feeds M2H-MX and RVIO2, which also uses IMU measurements. SuperPoint supports RVIO2 feature tracking, and the keypoint tracks also feed local graph refinement. Pose-aware temporal alignment uses RVIO2 poses to align recent M2H-MX depth and semantic predictions. The aligned depth supplies sparse depth factors to local graph refinement and, with the aligned labels, is integrated by volumetric fusion. Local graph refinement combines RVIO2 motion constraints, IMU constraints, and sparse depth factors, and removes keypoints on dynamic classes. It exports refined poses and loop-closure proposals to Hydra, which optimizes poses and mesh and updates the hierarchical scene graph (building, rooms, places, objects, and 3D mesh). This illustration was created from an author-prepared TikZ diagram using ChatGPT (OpenAI, GPT-6 Astra).}
  \label{fig:pipeline_overview}
\end{figure*}

As illustrated in Fig.~\ref{fig:pipeline_overview}, Mono-Hydra++ combines the
M2H-MX perception module with RVIO2 odometry, local graph refinement, and a
Mono-Hydra/Hydra mapping backend. RVIO2, with SuperPoint-guided tracking, supplies motion
estimates from monocular RGB and IMU. The local graph combines the resulting relative-motion constraints with visual
tracks, sparse predicted-depth factors, and semantic filtering, then exports refined
pose constraints to Hydra. Pose-aware temporal alignment uses camera motion to stabilize
depth and semantic observations before local graph refinement and volumetric fusion.

\subsection{M2H-MX Perception Interface}
\label{sec:perception_interface}
We use the M2H-MX model~\cite{udugama2026m2hmx} to map
an RGB image $I_t$ to a dense depth map $\hat{D}_t$ and semantic labels $\hat{S}_t$.
M2H-MX-B and M2H-MX-L share the decoder and task heads and differ in backbone capacity.
Only depth and semantics are consumed by the mapping pipeline; normals and edges are
auxiliary training outputs. We use the published M2H-MX architecture, adaptive-bin
depth head, and training objectives.

Depth is expressed in meters. We retain the indoor depth range $d_{\min}=0.1$\,m and
$d_{\max}=10.0$\,m and clamp predictions to $[d_{\min},d_{\max}]$ before pose-aware
temporal alignment (Sec.~\ref{sec:temporal_mapping}). The ScanNet training split and
exclusions are stated in Sec.~\ref{sec:scannet_prelim}.

\subsection{RVIO2 Odometry and Local Graph Refinement}
\label{sec:local_refinement}
RVIO2~\cite{huai2022square} estimates robocentric visual-inertial odometry. Its relative
pose estimates between keyframes enter a separate local graph containing keyframe
poses and tracked landmarks. This graph combines relative-pose, IMU, and visual
constraints with the sparse predicted-depth factors described below.
Optimization refines the local state and landmark estimates before pose constraints are
exported to Hydra. Depth-supported refinement therefore acts through this local graph;
it is not an additional residual in RVIO2's internal square-root update.
The following subsections describe feature tracking (Sec.~\ref{sec:sp_tracking}), metric
depth constraints on tracked landmarks (Sec.~\ref{sec:sparse_depth}), and the semantic
rejection of observations that
violate the static-scene assumption (Sec.~\ref{sec:semantic_masking}).
Sec.~\ref{sec:graph_interface} then describes how the refined poses are passed to Hydra.

\subsubsection{SuperPoint-Guided Feature Tracking}
\label{sec:sp_tracking}
The only change inside RVIO2 is its feature tracker (Fig.~\ref{fig:pipeline_overview}).
SuperPoint~\cite{detone2018superpoint} descriptors guide feature correspondence, and
photometric Lucas--Kanade (LK) alignment refines the matched locations to sub-pixel
accuracy. The IMU-propagated rotation first predicts the location
$\hat{\mathbf{k}}_i^t$ of track $i$ in image $I_t$. The track descriptor $\mathbf{d}_i$
is then compared with the dense SuperPoint descriptor field $\mathbf{f}_t$ in a window
$\Omega$ around this prediction, and LK alignment starts from the best match:
\begin{equation}
\boldsymbol{\delta}^\star=\arg\max_{\boldsymbol{\delta}\in\Omega}
\mathbf{d}_i^{\top}\mathbf{f}_t(\hat{\mathbf{k}}_i^t+\boldsymbol{\delta}),
\qquad
\mathbf{k}_i^t=\mathrm{LK}\!\left(I_{t-1},I_t,\mathbf{k}_i^{t-1};\,
\hat{\mathbf{k}}_i^t+\boldsymbol{\delta}^\star\right).
\label{eq:sp_tracking}
\end{equation}
A track is kept only if its descriptor peak is distinct from competing peaks and the
descriptor at the aligned position $\mathbf{k}_i^t$ still matches $\mathbf{d}_i$; the
descriptor is then updated. New tracks are initialized from SuperPoint keypoints.
Surviving tracks enter RVIO2's RANSAC and filter update. Aligned depth is sampled at the tracked keypoints
$\mathcal{K}_t=\{\mathbf{k}_i^t\}$, and semantic filtering removes observations
assigned to configured dynamic classes.

\subsubsection{Sparse Depth Factors with Motion Gating}
\label{sec:sparse_depth}
\begin{figure*}[!t]
\centering
\includegraphics[width=\textwidth]{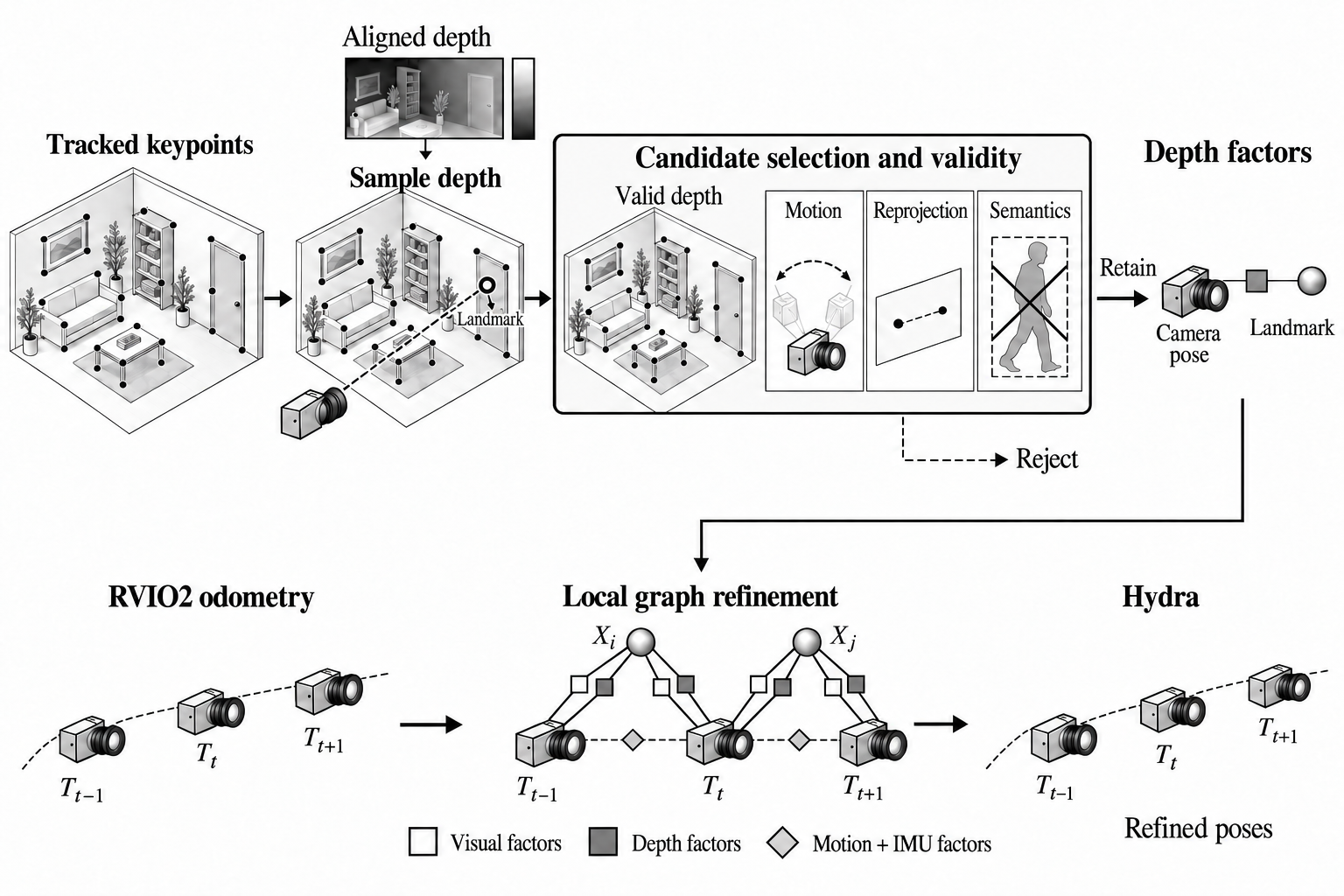}
\caption{Sparse depth factors. Aligned depth is sampled at tracked keypoints. Samples that pass the depth-validity, motion, reprojection, and semantic checks become depth factors between a camera pose and a landmark. These factors enter local graph refinement together with visual, RVIO2 motion, and IMU constraints, and the refined poses are passed to Hydra. This illustration was created from an author-prepared TikZ diagram using ChatGPT (OpenAI, GPT-6 Astra).}
\label{fig:depth_factors}
\end{figure*}

\begin{figure*}[!t]
\centering
\includegraphics[width=\textwidth]{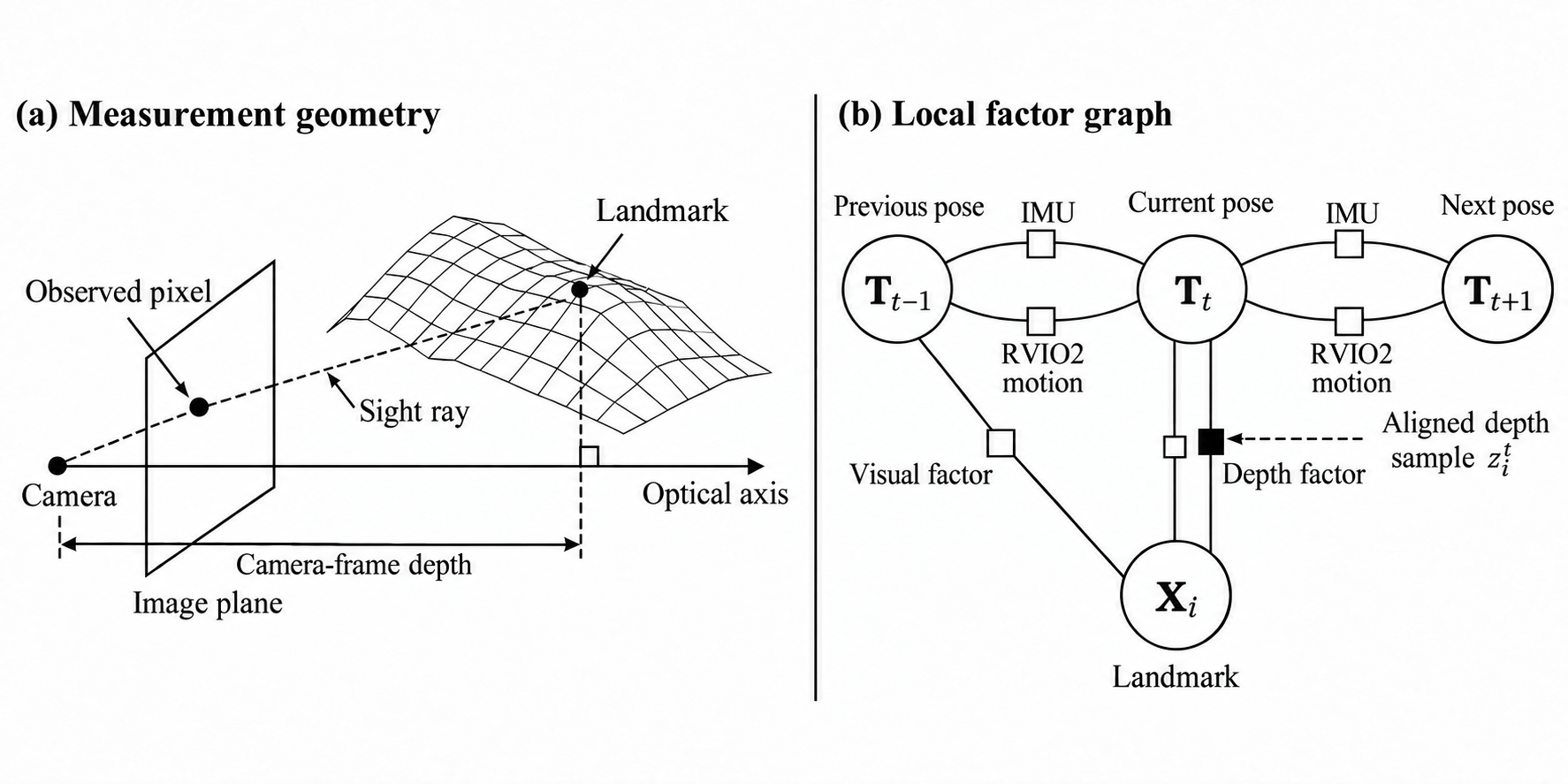}
\caption{Sparse depth factor geometry and local factor graph. (a) Camera-frame depth is the distance along the optical axis from the camera center to the projection of the landmark, which lies on the sight ray through the observed pixel. (b) Consecutive poses are linked by IMU and RVIO2 motion factors. A tracked landmark is constrained by visual factors and, at the current pose, by a depth factor using the aligned depth sample $z_i^t$. Circles denote variables and squares denote factors. Velocity and bias variables are omitted. This illustration was created from an author-prepared TikZ diagram using ChatGPT (OpenAI, GPT-6 Astra).}
\label{fig:depth_factor_graph}
\end{figure*}

A predicted-depth factor penalizes the difference between the depth predicted at a
tracked keypoint and the camera-frame depth of its associated 3D landmark. These sparse
factors constrain landmark depths in metric units during local graph refinement.
Fig.~\ref{fig:depth_factors}
summarizes how they are formed from the components of
Secs.~\ref{sec:sp_tracking}--\ref{sec:semantic_masking}: depth is sampled at the tracked
keypoints, the samples are checked, and only accepted samples become depth factors.
The samples are taken from the temporally aligned map $\bar{D}_t$
(Sec.~\ref{sec:temporal_alignment}). Each depth factor constrains an existing tracked
landmark, so we sample this dense map at its corresponding keypoint. The full depth map
is used separately for volumetric fusion. For each keypoint $\mathbf{k}_i^t$ we read a depth
measurement $z_i^t = \bar{D}_t(\mathbf{k}_i^t)$. Tracked keypoints with valid depth are
ranked using grayscale-image Sobel gradient magnitude, and up to 30 candidate
measurements per keyframe are retained. This favors well-textured, better-localized keypoints while
bounding the number of depth factors. Validity, motion, semantic, and reprojection checks
determine whether a retained point contributes to optimization; the number of accepted
factors can therefore be smaller than 30. Each accepted measurement adds a depth
factor that ties the landmark depth to the predicted depth along the ray:
\begin{equation}
r^{d}_{t,i} = z^{t}_{i} - \pi_z(\mathbf{T}_t \mathbf{X}_i), \quad
\mathcal{E}_{\mathrm{depth}} =
\sum_i \eta_{t,i}\rho\left(
\frac{r^{d}_{t,i}}{\sigma_d(z^{t}_{i})}
\right),
\qquad
\eta_{t,i}=g_{t,i}w_{t,i},
\label{eq:sparse_depth_factor}
\end{equation}
where $\mathbf{X}_i$ is the world-frame landmark associated with track $i$,
$\mathbf{T}_t$ maps it into the observing camera frame using the current pose and
calibrated camera--body extrinsics, and $\pi_z(\cdot)$ returns its camera-frame depth.
The Huber loss $\rho(\cdot)$ and depth-dependent noise model $\sigma_d(\cdot)$ control
the influence of uncertain observations. We use
\begin{equation}
\sigma_d(z)=\sigma_0+\sigma_1 z,\qquad
g_{t,i}=\mathbb{I}(\|\omega_t\|\leq\tau_\omega)\,
\mathbb{I}(\|\mathbf{e}^{\mathrm{repr}}_{t,i}\|_2\leq\tau_\pi),
\end{equation}
where $\sigma_0$ and $\sigma_1$ set the additive and depth-proportional uncertainty.
The term $g_{t,i}\in\{0,1\}$ is a binary acceptance gate: it equals one only if the
angular-velocity magnitude $\|\omega_t\|$ does not exceed the threshold $\tau_\omega$ and
the reprojection residual $\mathbf{e}^{\mathrm{repr}}_{t,i}$ does not exceed the pixel
threshold $\tau_\pi$; otherwise the factor is excluded. The final factor weight $\eta_{t,i}$ combines this
motion/reprojection gate with the semantic weight $w_{t,i}$ defined in Sec.~\ref{sec:semantic_masking}.
Motion gating disables depth factors under fast rotation, where image-to-depth
association is less reliable, and reprojection gating removes per-factor outliers that
remain geometrically inconsistent after projection. The constants
$\sigma_0,\sigma_1,\tau_\omega$, and $\tau_\pi$ are fixed for a reported
configuration rather than tuned per sequence. Retained depth
factors enter local graph refinement alongside the RVIO2-derived motion constraints.
Their effect reaches Hydra through the refined keyframe pose constraints.
Fig.~\ref{fig:depth_factor_graph} illustrates the camera-frame depth
measurement and its pose--landmark constraint.

\subsubsection{Semantic-Aware Robustification}
\label{sec:semantic_masking}
Dynamic objects corrupt geometric constraints in monocular VIO because their image
motion is not explained by the static-scene motion model. Semantic masking removes
keypoints and sparse depth factors on classes that are likely to violate the
static-scene assumption. In the reported experiments, the enabled dynamic mask is
$\mathcal{C}_{\mathrm{dyn}}=\{\text{person}\}$.
\begin{equation}
w_{t,i}=w(\mathbf{k}_i^t)=
\begin{cases}
0, & \text{if } \hat{S}_t(\mathbf{k}_i^t) \in \mathcal{C}_{\mathrm{dyn}},\\
1, & \text{otherwise},
\end{cases}
\label{eq:semantic_vio_mask}
\end{equation}
The semantic weight $w_{t,i}$ filters visual observations and the sparse depth factors
in Eq.~\eqref{eq:sparse_depth_factor} before local graph optimization. Thus, keypoints
assigned to dynamic classes do not supply visual or depth evidence to that refinement.

\subsubsection{Local-Graph-to-Hydra Interface}
\label{sec:graph_interface}

The refined local poses are exported as a keyframe pose graph. We use
$\mathbf{T}_{Wj}\in SE(3)$ to denote the pose that maps points from the body frame of
keyframe $j$ into the world frame $W$. Consecutive refined poses define odometry edges
between keyframes,
\begin{equation}
\mathbf{z}^{\mathrm{odom}}_{j,j+1}
=
(\hat{\mathbf{T}}_{Wj})^{-1}\hat{\mathbf{T}}_{W,j+1},
\qquad
\mathbf{e}^{\mathrm{odom}}_{j,j+1}
=
\mathrm{Log}\!\left(
(\mathbf{z}^{\mathrm{odom}}_{j,j+1})^{-1}
\mathbf{T}^{-1}_{Wj}\mathbf{T}_{W,j+1}
\right),
\label{eq:vio_odometry_edge}
\end{equation}
where $\hat{\mathbf{T}}_{Wj}$ and $\hat{\mathbf{T}}_{W,j+1}$ are consecutive refined
keyframe pose estimates, $\mathbf{T}^{-1}_{Wj}\mathbf{T}_{W,j+1}$ is the current
relative transform predicted by the backend pose variables, and $\mathrm{Log}(\cdot)$
maps the left-invariant pose residual to the Lie algebra.
The exported pose constraints are accompanied by loop-closure proposals, that is,
candidate constraints that link the current keyframe to an earlier keyframe at a
revisited location. Proposals come from the local graph and from Hydra, which detects
revisits using hierarchical scene-graph descriptors~\cite{hughes2022hydra}. Hydra
combines the incoming pose graph and accepted loop closures with mesh-deformation constraints and optimizes them jointly, so
that loop-closure corrections propagate to the metric-semantic mesh and layered scene
graph. The hierarchical
graph construction and global optimizer are inherited from Hydra.

\vspace{3mm}
\subsection{Pose-Aware Temporal Alignment and Mapping}
\label{sec:temporal_mapping}

\begin{figure*}[!t]
\centering
\includegraphics[width=\textwidth]{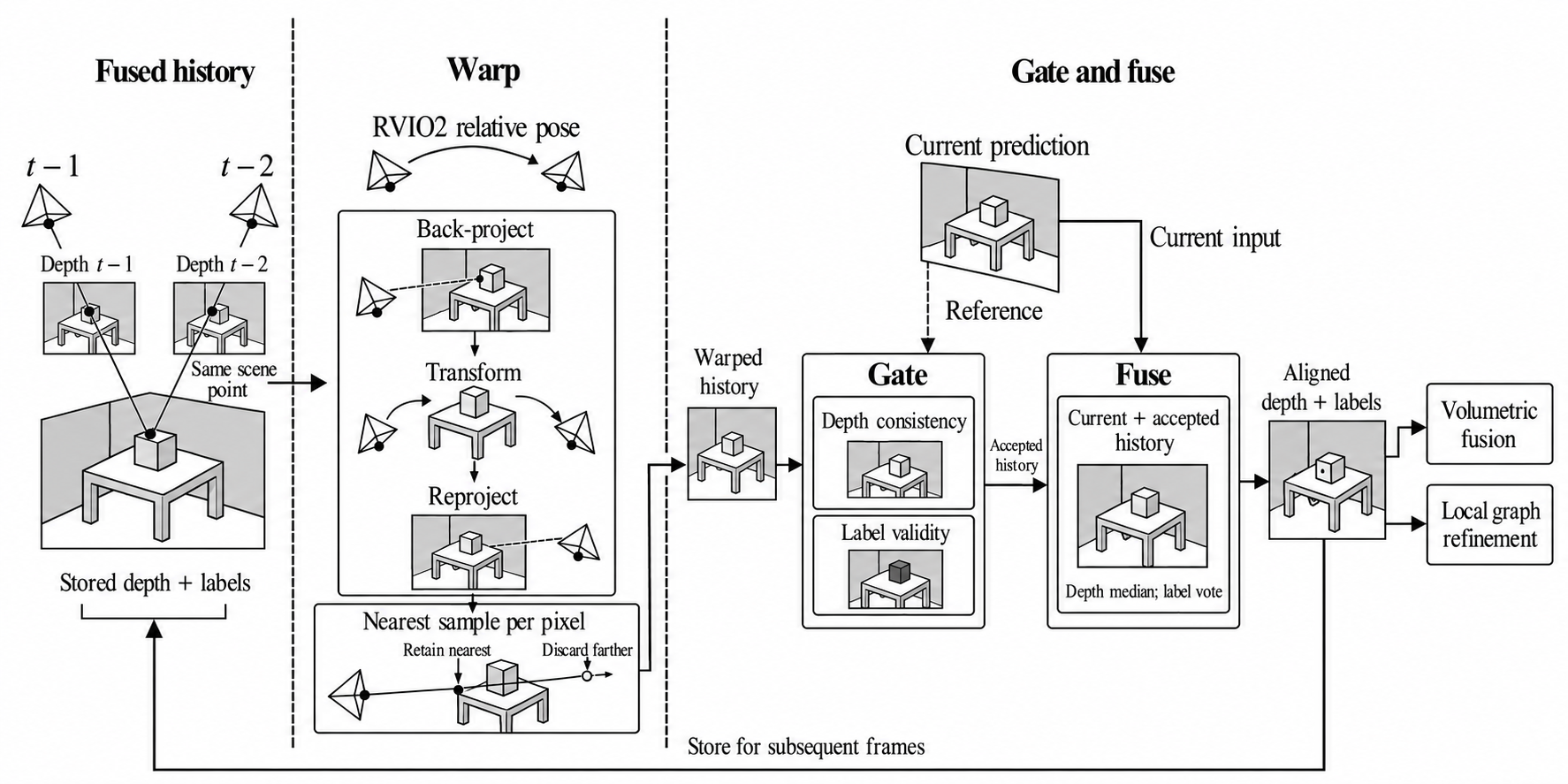}
\caption{Pose-aware temporal alignment. Previously fused depth and labels are warped into the current view using RVIO2 relative poses, keeping the nearest sample per pixel. Accepted history and the current prediction are combined by depth median and label voting, with depth-consistency checks and label exclusions applied to their respective streams. The fused outputs supply local graph refinement and volumetric fusion and are stored for subsequent frames. The current prediction is retained where historical support is insufficient. This illustration was created from an author-prepared TikZ diagram using ChatGPT (OpenAI, GPT-6 Astra).}
\label{fig:temporal_alignment}
\end{figure*}

Pose-aware temporal alignment uses VIO poses to warp previously fused depth and semantic
maps into the current camera frame. It combines accepted history with current predictions
to reduce temporal flicker before local graph refinement and volumetric fusion.
Volumetric fusion integrates these observations into a shared 3D
voxel map: Hydra updates a truncated signed distance field (TSDF) for geometry and
accumulates semantic evidence in the volume.
Fig.~\ref{fig:temporal_alignment} summarizes the geometric warp, validity checks,
and fallback used to prepare these observations.

\subsubsection{Temporal Pose-Warp Alignment}
\label{sec:temporal_alignment}
Even with strong per-frame predictions, monocular depth and semantic outputs can flicker
across frames, which degrades fusion and produces ghosting in the map. To reduce this,
we retain up to $K$ previously fused depth and semantic maps and warp them into the
current camera frame using VIO poses and intrinsics. The accepted historical samples
are combined with the current predictions. The resulting maps are published and stored
for subsequent frames; each stored map can therefore contain information from earlier
frames. This temporal correction is applied after M2H-MX inference.
We denote the retained past-frame indices by $\mathcal{T}_K\subseteq\{1,\ldots,K\}$.
The reported temporal ablation uses
$K\in\{0,1,3,5\}$, where $K=0$ disables temporal fusion. For each past frame $t-\tau$, we
back-project,
transform, and reproject:
\begin{equation}
\label{eq:temporal_pose_warp}
\mathbf{P}_{t-\tau}(\mathbf{x}) = \Pi^{-1}(\mathbf{x}, \bar{D}_{t-\tau}(\mathbf{x})),
\qquad
\tilde{\mathbf{P}}_{t-\tau \rightarrow t} =
\mathbf{T}_{t-\tau \rightarrow t}\,\mathbf{P}_{t-\tau},
\end{equation}
\begin{equation}
\label{eq:temporal_reprojection}
\tilde{\mathbf{x}} = \Pi(\tilde{\mathbf{P}}_{t-\tau \rightarrow t}), \qquad
\tilde{D}_{t-\tau \rightarrow t}(\tilde{\mathbf{x}}) =
\pi_z(\tilde{\mathbf{P}}_{t-\tau \rightarrow t}).
\end{equation}
Reprojected samples are z-buffered in the current frame, and only the nearest valid
sample per pixel is retained before temporal fusion. The stored fused labels
$\bar{S}_{t-\tau}$ are warped with nearest-neighbor sampling to obtain
$\tilde{S}_{t-\tau \rightarrow t}$. Let $v_t$ and $v_{t-\tau\rightarrow t}$ indicate
valid current and warped depths, respectively. We accept a warped depth if it agrees
with the current depth, or if the current depth is invalid:
\begin{equation}
\label{eq:temporal_gate}
\alpha_{t-\tau}(\mathbf{x}) =
v_{t-\tau\rightarrow t}(\mathbf{x})\,
\mathbb{I}\big(v_t(\mathbf{x})=0\;\lor\;
|\tilde{D}_{t-\tau\rightarrow t}(\mathbf{x})-\hat{D}_t(\mathbf{x})|
\leq\delta_d(\mathbf{x})\big),
\end{equation}
where $\delta_d(\mathbf{x})=\max(\delta_{\mathrm{abs}},
\delta_{\mathrm{rel}}\hat{D}_t(\mathbf{x}))$ combines absolute and relative depth
tolerances. Let $\mathcal{Z}_t(\mathbf{x})$ be the multiset containing the valid current
depth and each warped depth accepted by $\alpha_{t-\tau}$. Depth fusion takes their
median:
\begin{equation}
\bar{D}_t(\mathbf{x}) =
\operatorname{median}\big(\mathcal{Z}_t(\mathbf{x})\big).
\end{equation}
Labels are voted on separately. Let $\beta_t$ indicate a valid current depth and a
non-excluded current label, and let $\beta_{t-\tau}$ combine the depth gate
$\alpha_{t-\tau}$ with the same label-validity check for warped labels. Configured
dynamic and ignored classes are excluded from voting. The support for class $c$ is
\begin{equation}
V_t(c,\mathbf{x}) = \beta_t(\mathbf{x})\,
\mathbb{I}\big(\hat{S}_t(\mathbf{x})=c\big)
+\sum_{\tau\in\mathcal{T}_K}\beta_{t-\tau}(\mathbf{x})\,
\mathbb{I}\big(\tilde{S}_{t-\tau\rightarrow t}(\mathbf{x})=c\big).
\end{equation}
The winning class replaces the current label only if its support is at least two and
strictly exceeds the current label's support. Ties and current labels belonging to
excluded classes retain $\hat{S}_t$. Without accepted history, the current prediction
is retained. The fused pair $(\bar{D}_t,\bar{S}_t)$ is then inserted into the history
buffer. The ablation setting $K=0$ bypasses the filter and uses the current prediction.
The fused maps combine pose-consistent evidence across frames before local graph
refinement and volumetric fusion.
\section{Experiments}
\label{sec:experiments}

We measure trajectory error, reconstruction accuracy, and object-node and scene-graph
quality. The ablations compare sparse depth factors, semantic masking, and temporal
window lengths; the deployment study reports mapping accuracy and processing times.
Standalone dense-prediction benchmarks are reported in the M2H-MX
paper~\cite{udugama2026m2hmx}.

\subsection{Datasets}
\label{sec:datasets}

ScanNet~\cite{dai2017scannet} is used for system-level trajectory, semantic mesh, and
object-level evaluation; 7-Scenes~\cite{shotton2013scene} is used for calibrated
trajectory and reconstruction benchmarking; uHumans2~\cite{rosinol2021kimera} is used
for dynamic-scene ablations; and the ITC real-world
dataset~\cite{udugama2023monohydra} is used for deployment-oriented indoor mapping
evaluation. For ScanNet system-level evaluation,
input frames and sequence metadata are read from the released \texttt{.sens} files.
Mono-Hydra++ uses the RGB stream; ScanNet depth frames are not used as input to the
proposed pipeline. For ScanNet and 7-Scenes, IMU measurements were simulated from the
provided camera trajectories following the simulation approach of
OpenVINS~\cite{geneva2020openvins}.

Model training was performed on an NVIDIA A40 GPU. Real-time inference and system-level
results for M2H-MX-L were measured on an NVIDIA RTX 4080 Super 16GB.
Unless stated otherwise, the benchmark results use sparse depth factors, semantic
masking, and temporal alignment with $K=3$, the best trajectory setting in the ablation
(Sec.~\ref{sec:uhumans2_ablation}).

\subsection{Evaluation Metrics}
\label{sec:metrics}

Camera trajectory quality is summarized by absolute trajectory error (ATE). Mono-Hydra++
trajectories are SE(3)-aligned to the ground truth without scale correction. On
7-Scenes, reconstruction quality is measured by accuracy, completeness, and Chamfer
distance. For ScanNet semantic-mesh evaluation, we retain three metrics: global
mesh-vertex mIoU, Radius F1@0.5m, and Box F1@0.25. In Radius F1@0.5m, a prediction is
counted as correct when its center lies within 0.5 m of the ground truth. Box F1@0.25
uses one-to-one same-class matching between predicted and ground-truth 3D object boxes,
with a true positive requiring 3D bounding-box IoU $\geq 0.25$. Unmatched predictions
are false positives and unmatched ground-truth objects are false negatives. Higher
values are better for mIoU and F1, while lower values are better for ATE,
accuracy, completeness, and Chamfer distance.

For scene-graph evaluation, Node F1 measures object-node detection quality after
matching predicted and ground-truth object nodes using class consistency and the spatial
matching criterion used for the corresponding experiment. Scene graph similarity is computed as a
normalized graph-edit similarity,
\begin{equation}
S_{\mathrm{SG}} =
1-\frac{d_{\mathrm{SG}}(\hat{G},G)}
{|\hat{V}|+|V|+|\hat{E}|+|E|+\epsilon},
\qquad
d_{\mathrm{SG}}=d_V(\hat{V},V)+d_E(\hat{E},E),
\end{equation}
where $G=(V,E)$ is the reference scene graph, $\hat{G}=(\hat{V},\hat{E})$ is the
predicted graph, $d_V$ counts class-aware node insertions, deletions, and substitutions
after spatial matching, and $d_E$ counts relation-edge insertions, deletions, and
substitutions over the matched nodes. The node and edge edit costs are equally weighted;
higher values therefore indicate closer agreement in both graph structure and
semantic-spatial assignment.

\FloatBarrier

\begin{table}[!t]
\centering
\footnotesize
\setlength{\tabcolsep}{4pt}
\renewcommand{\arraystretch}{1.05}
\caption{ATE (cm) on the ScanNet sequences following the Go-SLAM selected-sequence
evaluation protocol. DROID-SLAM (VO) excludes global bundle adjustment; iMAP and
NICE-SLAM results are from their papers. Best results are in bold; second best are
underlined.}
\label{tab:scannet_scenes}
\resizebox{\textwidth}{!}{%
\begin{tabular}{llrrrrrrrrr}
\toprule
Input & Method &
0000 & 0054 & 0059 & 0106 & 0169 & 0181 & 0233 & 0465 & Avg. \\
\midrule
\multicolumn{2}{c}{\# Frames} & 5578 & 6629 & 1807 & 2324 & 2034 & 2349 & 7643 & 6306
& -- \\
\midrule
\multirow{5}{*}{RGB-D}
& iMAP~\cite{Sucar:etal:ICCV2021} & 55.95 & 70.11 & 32.06 & 17.50 & 70.51 & 32.10 &
  86.42 & 85.03 & 56.21 \\
& NICE-SLAM~\cite{Zhu2022CVPR} & 8.64 & 20.93 & 12.25 & 8.09 & 10.28 & 12.93 & 9.00 &
  22.31 & 13.05 \\
& DROID-SLAM (VO)~\cite{teed2021droid} & 8.00 & 29.28 & 11.30 & 9.97 & 8.64 & 7.38 &
  6.75 & 11.37 & 11.59 \\
& DROID-SLAM~\cite{teed2021droid} & 5.36 & 8.89 & 7.72 & 7.06 & 8.01 & 6.97 & 4.90 &
  \underline{8.32} & 7.15 \\
& Go-SLAM~\cite{zhang2023goslam} & \underline{5.35} & \underline{8.75} &
  \underline{7.52} & 7.03 & \underline{7.74} & \underline{6.84} &
  \underline{4.78} & \textbf{8.15} & \underline{7.02} \\
\midrule
\multirow{5}{*}{Mono}
& ORB-SLAM3~\cite{campos2021orb} & 73.93 & 243.26 & 90.67 & 178.13 & 60.15 & 104.93 &
  25.01 & 181.86 & 119.74 \\
& DROID-SLAM (VO)~\cite{teed2021droid} & 11.05 & 204.31 & 67.26 & 11.20 & 16.21 & 9.94
  & 71.08 & 117.84 & 63.61 \\
& DROID-SLAM~\cite{teed2021droid} & 5.48 & 197.71 & 9.00 & \underline{6.76} & 7.86 &
  7.41 & 72.23 & 114.36 & 52.60 \\
& Go-SLAM~\cite{zhang2023goslam} & 5.94 & 13.29 & 8.27 & 8.07 & 8.42 & 8.29 & 5.31 &
  79.51 & 17.14 \\
& Mono-Hydra++ & \textbf{5.08} & \textbf{8.69} & \textbf{7.14} &
  \textbf{6.75} & \textbf{7.28} & \textbf{6.30} & \textbf{4.63} & 9.38 &
  \textbf{6.91} \\
\bottomrule
\end{tabular}}
\end{table}

\FloatBarrier

\subsection{ScanNet Scene-Level Comparison}
\label{sec:scannet_prelim}

Table~\ref{tab:scannet_scenes} reports ATE on selected ScanNet sequences following the
Go-SLAM evaluation protocol, for RGB-D and monocular baselines. DROID-SLAM (VO) denotes
the visual odometry variant
without global bundle adjustment; iMAP and NICE-SLAM results are reported from their
respective papers.
The Mono-Hydra++ results in Table~\ref{tab:scannet_scenes} use M2H-MX weights trained
on ScanNet-25k, a subset of ScanNet's 2D depth and semantic annotations. The selected
ScanNet evaluation sequences were excluded from the M2H-MX training split.

\begin{figure}[p]
\centering
{\normalsize\bfseries ScanNet RGB-D\par}
\vspace{6pt}
\noindent
\qualcell{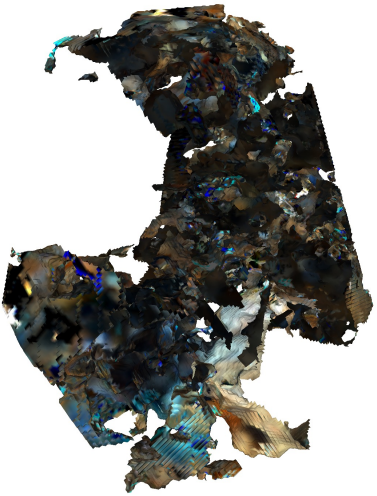}{ATE: 70.11 cm}{iMAP~\cite{Sucar:etal:ICCV2021}}\hfill
\qualcell{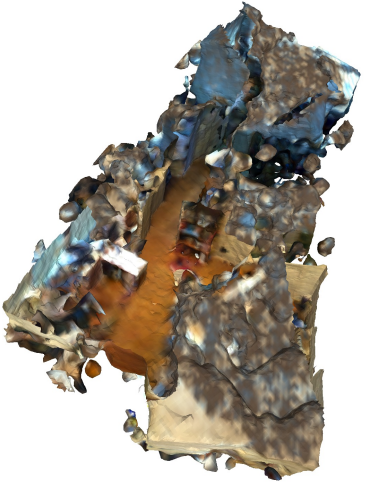}{ATE: 20.93 cm}
    {NICE-SLAM~\cite{Zhu2022CVPR}}\hfill
\qualcell{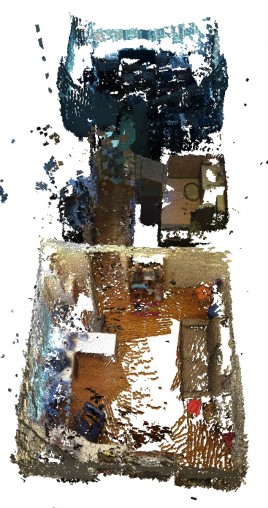}{ATE: 8.89 cm}
    {DROID-SLAM~\cite{teed2021droid}}\hfill
\qualcell{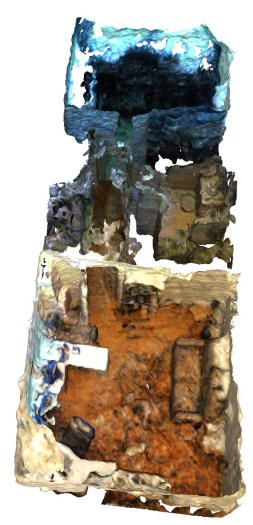}{ATE: 8.75 cm}
    {Go-SLAM~\cite{zhang2023goslam}}\par
\vspace{14pt}
{\normalsize\bfseries ScanNet Mono\par}
\vspace{6pt}
\noindent
\qualcell{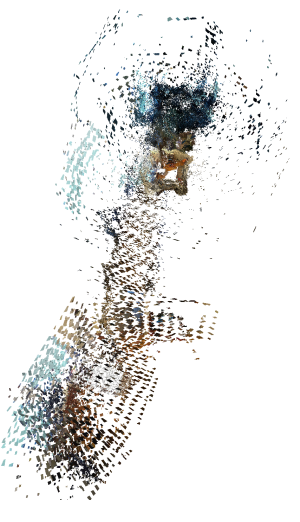}{ATE: 197.71 cm}
    {DROID-SLAM~\cite{teed2021droid}}\hfill
\qualcell{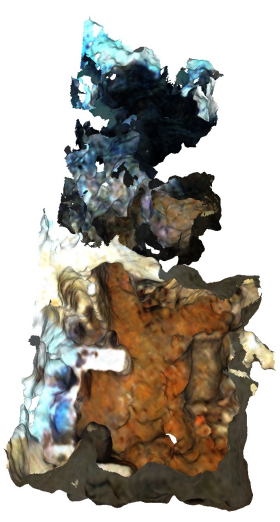}{ATE: 13.29 cm}
    {Go-SLAM~\cite{zhang2023goslam}}\hfill
\qualcell{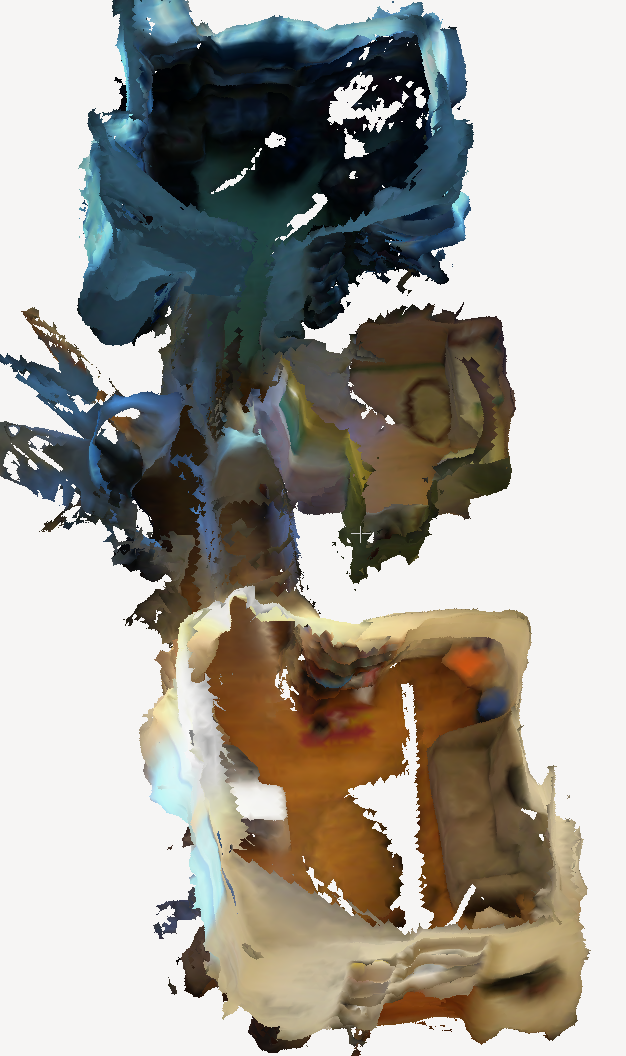}{ATE: 8.69 cm}{Mono-Hydra++}\hfill
\qualcell{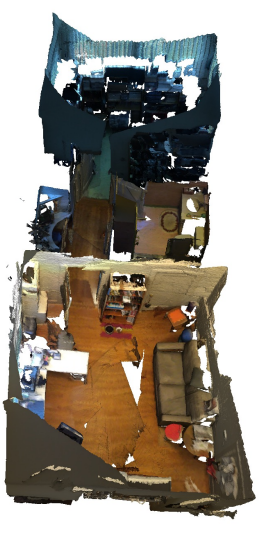}{Ground truth}{GT}\par
\vspace{6pt}
\caption{Qualitative comparison on ScanNet scene0054. RGB-D results are on the top
row, monocular results are on the bottom row, and ground truth is the rightmost panel.}
\label{fig:scannet_qual}
\end{figure}

Table~\ref{tab:scannet_scenes} shows that Mono-Hydra++ attains the lowest average ATE
among the monocular methods and the best score on seven of the eight reported
sequences. Among the RGB-D baselines, Go-SLAM yields the lowest average ATE and the
best scene0465 result. Fig.~\ref{fig:scannet_qual} shows the
corresponding qualitative reconstruction on scene0054. On the selected ScanNet
sequences, Mono-Hydra++ achieves an average ATE of 6.91\,cm, 1.6\% lower than RGB-D
Go-SLAM (7.02\,cm), without RGB-D depth or LiDAR input.

\FloatBarrier

\subsection{7-Scenes ATE Benchmark}
\label{sec:7scenes_ate}
Table~\ref{tab:7scenes_ate} reports ATE RMSE on 7-Scenes. We include benchmark rows from
the VGGT-SLAM comparison table~\cite{maggio2025vggt-slam} together with the
Mono-Hydra++ result using calibrated intrinsics.

\begin{table}[!t]
\centering
\scriptsize
\setlength{\tabcolsep}{3pt}
\caption{ATE RMSE (m) on 7-Scenes (lower is better). Rows under ``Calib.'' use calibrated
camera intrinsics, while rows under ``Uncalib.'' follow the uncalibrated-sequence
protocol.}
\label{tab:7scenes_ate}
\resizebox{\textwidth}{!}{%
\begin{tabular}{llrrrrrrrr}
\toprule
Protocol & Method & chess & fire & heads & office & pumpkin & kitchen & stairs & Avg \\
\midrule
\multirow{4}{*}{Uncalib.}
& DROID-SLAM~\cite{teed2021droid} & 0.047 & 0.038 & 0.034 & 0.136 & 0.166 & 0.080 & 0.044 & 0.078 \\
& MASt3R-SLAM~\cite{murai2024_mast3rslam} & 0.063 & 0.046 & 0.029 & 0.103 & 0.114 & 0.074 & 0.032 & 0.066 \\
& VGGT-SLAM~\cite{maggio2025vggt-slam} (Sim(3), $w=32$) & 0.037 & 0.026 & \underline{0.018} & 0.104 & 0.133 & 0.061 & 0.093 & 0.067 \\
& VGGT-SLAM~\cite{maggio2025vggt-slam} (SL(4), $w=32$) & 0.036 & 0.028 & \underline{0.018} & 0.103 & 0.133 & 0.058 & 0.093 & 0.067 \\
\midrule
\multirow{4}{*}{Calib.}
& NICER-SLAM~\cite{zhu2024nicer} & \underline{0.033} & 0.069 & 0.042 & 0.108 & 0.200 & \underline{0.039} & 0.108 & 0.086 \\
& DROID-SLAM~\cite{teed2021droid} & 0.036 & 0.027 & 0.025 & \underline{0.066} & 0.127 & 0.040 & \underline{0.026} & 0.049 \\
& MASt3R-SLAM~\cite{murai2024_mast3rslam} & 0.053 & \underline{0.025} & \textbf{0.015} & 0.097 & \underline{0.088} & 0.041 & \textbf{0.011} & \underline{0.047} \\
& Mono-Hydra++ (ScanNet wts) & \textbf{0.018} & \textbf{0.019} & 0.031 & \textbf{0.034} & \textbf{0.024} & \textbf{0.037} & 0.071 & \textbf{0.033} \\
\bottomrule
\end{tabular}}
\end{table}

The uncalibrated VGGT-SLAM rows and the calibrated Mono-Hydra++ row use different
protocols and are interpreted separately. Within the calibrated setting,
Mono-Hydra++ attains the best average ATE and the best scores on chess, fire, office,
pumpkin, and kitchen. It is evaluated with ScanNet-trained weights, without
7-Scenes-specific fine-tuning because 7-Scenes does not provide the semantic
supervision required by our multi-task model.
These results compare complete trained systems within each calibration setting.
Training data and architectures differ: DROID-SLAM reports
training on synthetic TartanAir~\cite{teed2021droid}, while MASt3R-SLAM and VGGT-SLAM
use pretrained geometry networks~\cite{murai2024_mast3rslam,maggio2025vggt-slam}.

\subsection{7-Scenes Reconstruction Quality}
Table~\ref{tab:7scenes_recon} reports 7-Scenes reconstruction quality using the same
metric convention as the VGGT-SLAM benchmark~\cite{maggio2025vggt-slam}. As in the ATE
table, uncalibrated rows are shown first and calibrated rows are listed below.

\begin{table}[!t]
\centering
\scriptsize
\setlength{\tabcolsep}{3pt}
\caption{7-Scenes reconstruction quality, RMSE (m; lower is better).}
\label{tab:7scenes_recon}
\begin{tabular}{llcccc}
\toprule
Protocol & Method & ATE $\downarrow$ & Acc. $\downarrow$ & Complet. $\downarrow$ & Chamfer $\downarrow$ \\
\midrule
\multirow{3}{*}{Uncalib.}
& MASt3R-SLAM~\cite{murai2024_mast3rslam} & 0.066 & \underline{0.068} & \textbf{0.045} & \underline{0.056} \\
& VGGT-SLAM~\cite{maggio2025vggt-slam} (Sim(3), $w=32$) & 0.067 & \textbf{0.052} & 0.062 & 0.057 \\
& VGGT-SLAM~\cite{maggio2025vggt-slam} (SL(4), $w=32$) & 0.067 & \textbf{0.052} & 0.058 & \textbf{0.055} \\
\midrule
\multirow{3}{*}{Calib.}
& DROID-SLAM~\cite{teed2021droid} & 0.049 & 0.141 & \underline{0.048} & 0.094 \\
& MASt3R-SLAM~\cite{murai2024_mast3rslam} & \underline{0.047} & 0.089 & 0.085 & 0.087 \\
& Mono-Hydra++ (ScanNet wts) & \textbf{0.033} & 0.104 & 0.058 & 0.086 \\
\bottomrule
\end{tabular}
\end{table}

Within the calibrated protocol, Mono-Hydra++ achieves the best ATE and the lowest
Chamfer distance (0.086\,m compared with 0.087\,m for MASt3R-SLAM), while MASt3R-SLAM yields the strongest calibrated accuracy and
DROID-SLAM the strongest calibrated completeness.

\subsection{3D Semantic Mesh Evaluation}
\label{sec:scannet_sem_mesh}
We evaluate semantic mesh quality on ScanNet-v2 scenes by transferring predicted mesh
labels onto ScanNet ground-truth mesh vertices and computing IoU on ScanNet20 classes.
All compared perception models are run within the same Mono-Hydra++ pipeline, so the
mapping and scene-graph stages are common to every result and only the multi-task
perception model differs. This comparison therefore isolates the effect of perception on
downstream mapping quality; the pipeline components themselves are assessed in the
ablations of Sec.~\ref{sec:uhumans2_ablation}. First, predicted and GT meshes
are rigidly aligned (ICP-based alignment). Then, for each GT vertex, we assign the label
of its nearest predicted mesh vertex (KD-tree nearest neighbor). Ground-truth unlabeled
vertices are ignored.

Global mIoU, Radius F1@0.5m, and Box F1@0.25 follow the definitions in
Sec.~\ref{sec:metrics}. In the object-level evaluations, ``Missing GT'' denotes a ground-truth object instance for
which no same-class prediction satisfies the matching criterion, either within the
radius threshold or at the required box IoU. It is therefore counted as a false
negative rather than a missing annotation.

\begin{figure}[!t]
\centering
\includegraphics[width=0.86\textwidth]{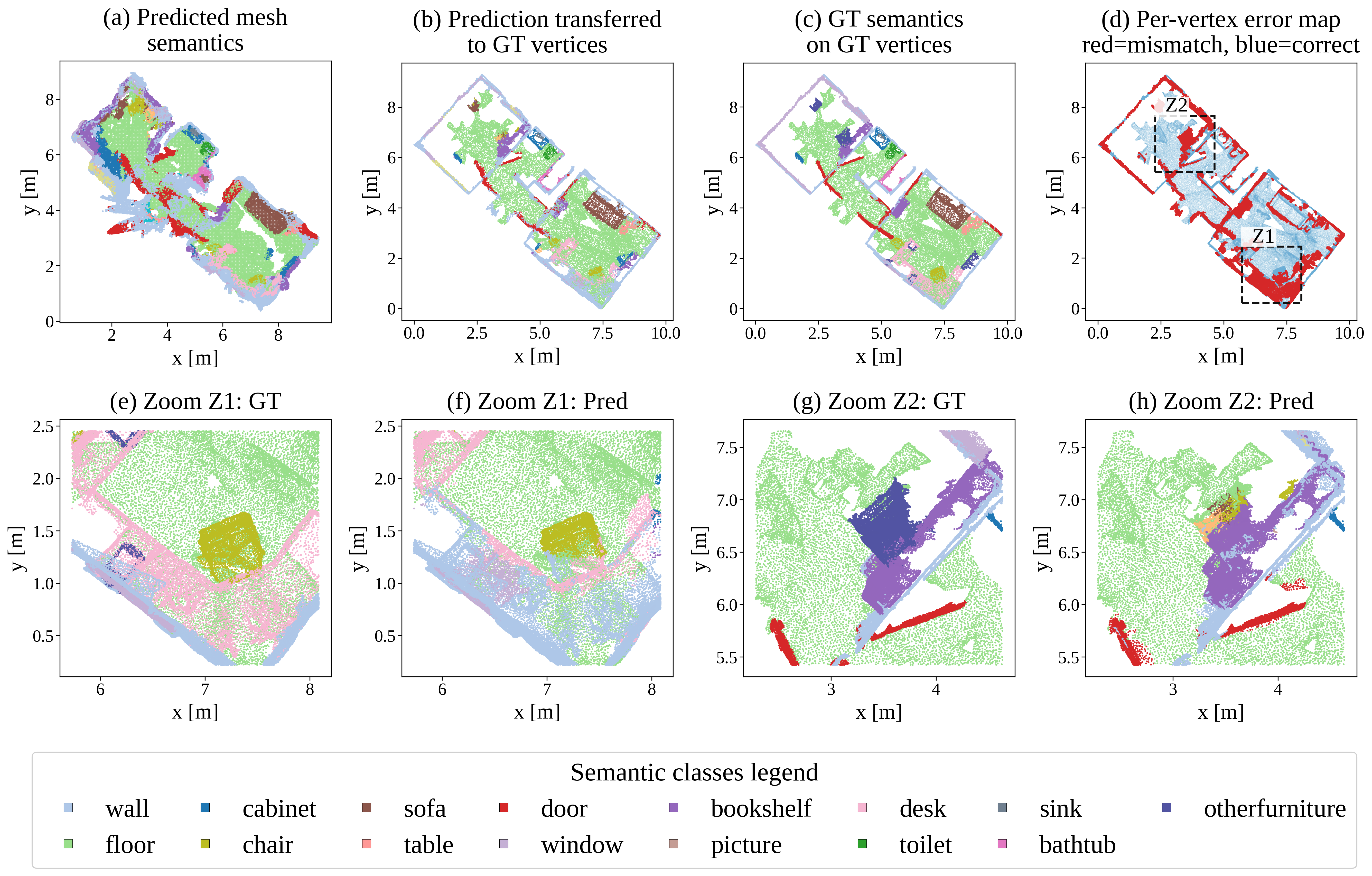}
\\[-1mm]
{\small (a) ScanNet scene0054}

\vspace{0.5mm}
\includegraphics[width=0.86\textwidth]{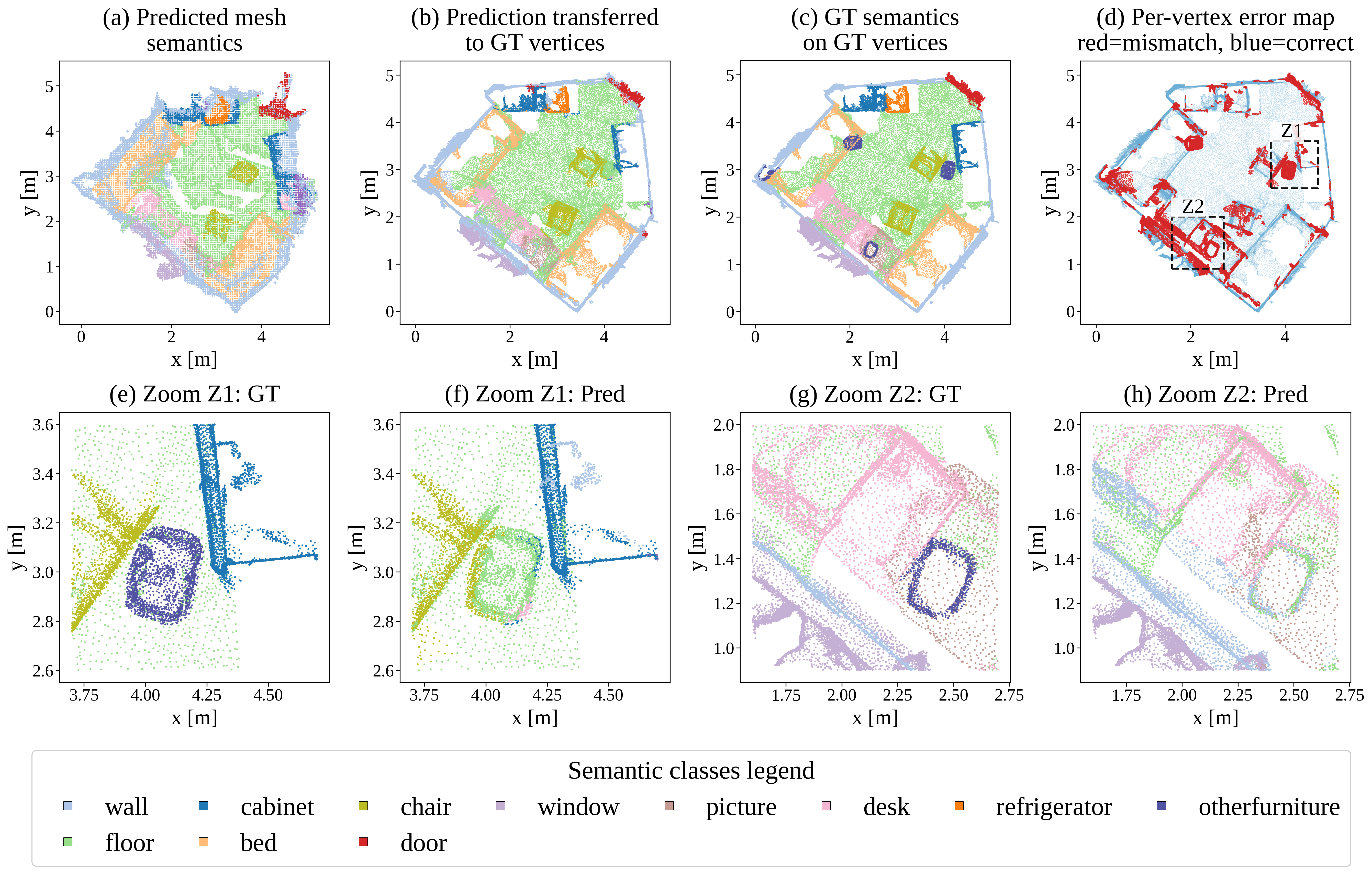}
\\[-1mm]
{\small (b) ScanNet scene0233}
\caption{Qualitative semantic mesh comparison on two representative ScanNet scenes. For
each scene, the top row shows the predicted semantic mesh, the predicted labels
transferred onto GT vertices, the GT semantic mesh, and the per-vertex mismatch map;
the bottom row shows zoomed failure regions. Errors concentrate near object boundaries,
thin structures, cluttered furniture regions, and, in scene0233, confusion involving
the broad otherfurniture category.}
\label{fig:semantic_mesh_qual}
\end{figure}

\begin{figure}[!t]
\centering
\includegraphics[width=0.84\textwidth]{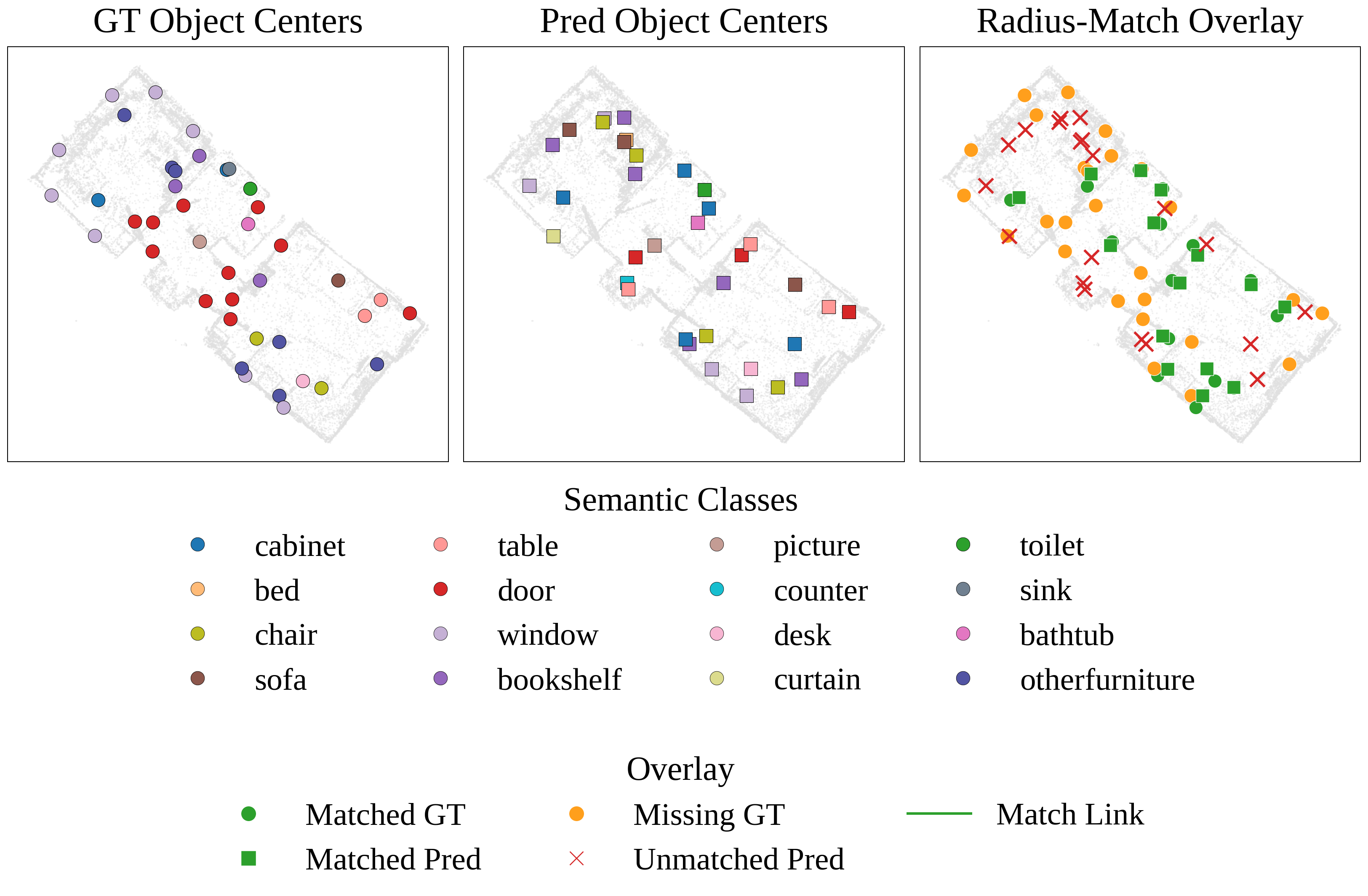}
\\[-1mm]
{\footnotesize (a) ScanNet scene0054}

\vspace{0.5mm}
\includegraphics[width=0.84\textwidth]{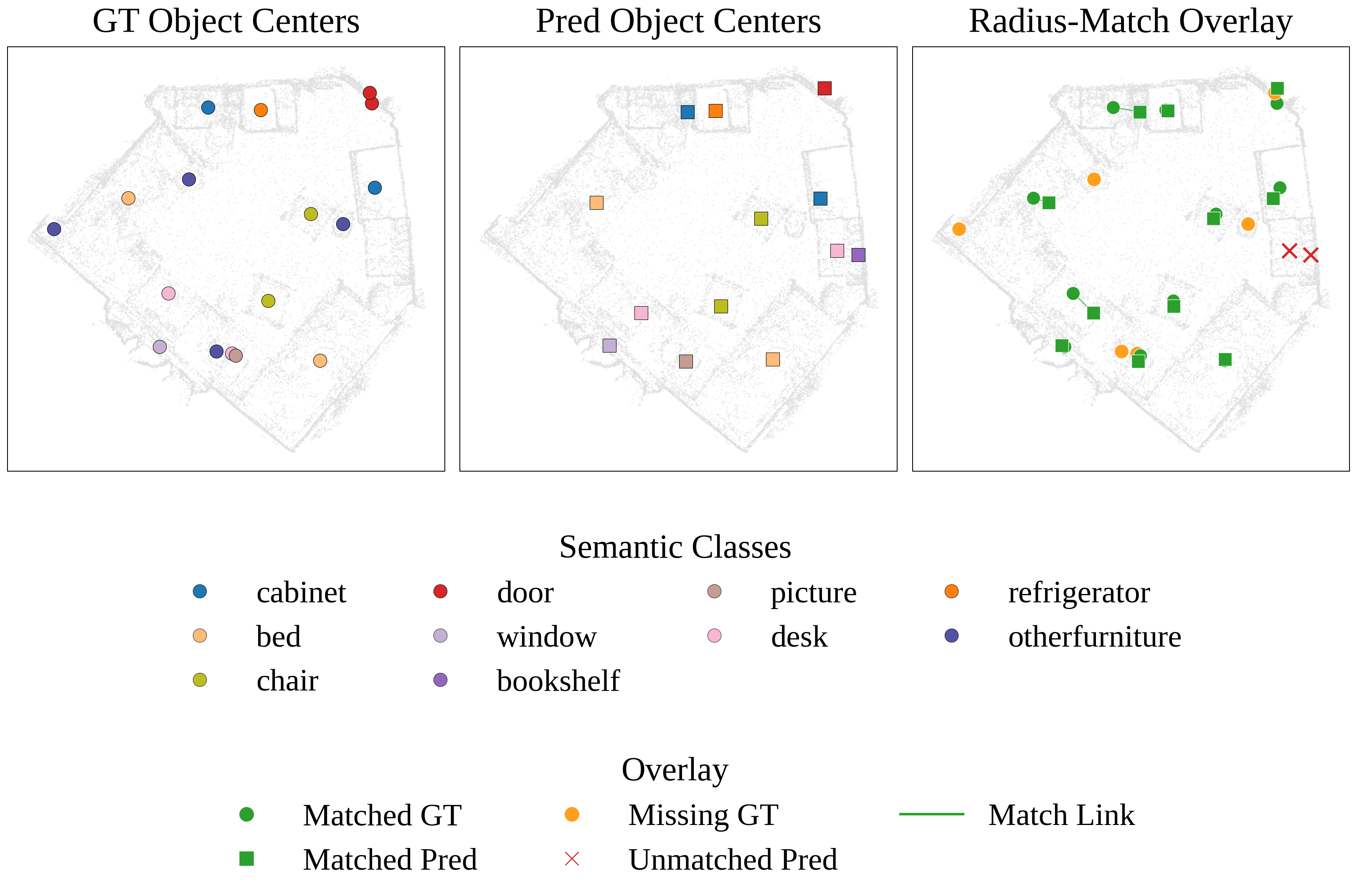}
\\[-1mm]
{\footnotesize (b) ScanNet scene0233}
\caption{Qualitative radius-based object retrieval at $r=0.5$\,m on the same two
ScanNet scenes used in Fig.~\ref{fig:semantic_mesh_qual}. Each row shows ground-truth
object centers, predicted object centers, and the matching overlay. Green connections
indicate same-class matches within the radius threshold; missing GT and unmatched
predictions highlight object-level failures in cluttered regions.}
\label{fig:radius_presence_qual}
\end{figure}

To better contextualize these results, we compare against recent 3D semantic
segmentation methods that assume reconstructed 3D geometry or dense point clouds and
should therefore be read as semantic upper bounds rather than matched baselines for an
online monocular SLAM pipeline. Point Transformer V3~\cite{wu2024ptv3} reports 78.6
validation mIoU and 79.4 test mIoU on ScanNet, while DINO in the
Room~\cite{knaebel2026ditr} reports 80.5 validation mIoU and 79.7 test mIoU. Figs.~\ref{fig:semantic_mesh_qual}
and~\ref{fig:radius_presence_qual} provide qualitative semantic-mesh and object-level
results on representative ScanNet scenes.

Figure~\ref{fig:semantic_mesh_qual} provides a qualitative view of semantic mesh
performance beyond the aggregate mIoU scores. In both representative scenes, the
predicted mesh preserves the overall room geometry and recovers the dominant semantic
layout with good spatial coherence. Large planar structures such as walls and floors
remain well aligned, and the main furniture arrangement is broadly consistent with the
scene topology. The remaining errors are concentrated around object boundaries, thin
structures, and cluttered furniture areas where partial observations, geometric
fragmentation, or category ambiguity are more pronounced. The zoomed regions indicate
that many failures correspond to semantic leakage between adjacent categories rather
than gross geometric misalignment. In particular, some ambiguous object regions are
absorbed into neighboring dominant classes, and the otherfurniture category is not
preserved reliably as a distinct label, which is especially visible in the scene0233
bottom-row zoom-ins.

Table~\ref{tab:overall_quant_results} summarizes the unified semantic-mesh and
object-level evaluation using the three primary metrics above, with each multi-task
perception model evaluated inside the same Mono-Hydra++ pipeline and under the same
protocol.

\begin{table}[!t]
\centering
\caption{Quantitative comparison on ScanNet. We report global mesh-vertex mIoU for
semantic quality, Radius F1@0.5m for object-level localization, and Box F1@0.25 for
object bounding accuracy. All perception models are evaluated within the same
Mono-Hydra++ pipeline.}
\label{tab:overall_quant_results}
\setlength{\tabcolsep}{5pt}
\resizebox{\textwidth}{!}{%
\begin{tabular}{lccc}
\toprule
Method in Mono-Hydra++ & Global mIoU $\uparrow$ & Radius F1@0.5m $\uparrow$ & Box F1@0.25 $\uparrow$ \\
\midrule
MTMamba++ & 28.38 & 34.57 & 23.70 \\
M2H & 41.42 & 35.93 & 29.95 \\
M2H-MX & 44.96 & 42.59 & 33.81 \\
\bottomrule
\end{tabular}
}
\end{table}

Mono-Hydra++ with M2H-MX achieves 44.96 global mIoU, 42.59 Radius F1@0.5m, and
33.81 Box F1@0.25 on the evaluated sequences, outperforming the MTMamba++ and M2H
variants across all three metrics. Because the pipeline is otherwise unchanged, these
gains are consistent with improved downstream semantic mesh quality, object-level
localization, and box accuracy from the M2H-MX perception model.

The pooled confusion analysis also reveals a systematic failure mode beyond isolated
class swaps. Object regions are frequently absorbed into dominant structural classes,
particularly wall and floor, rather than only being confused with semantically adjacent
object classes. This is most evident for otherfurniture, which attains zero IoU over
the evaluated scenes and is reassigned primarily to wall (39.3\%), floor (19.3\%), and
bookshelf (14.8\%). Chair errors are likewise dominated by wall (49.1\% of chair false
negatives), table (22.5\%), and floor (13.7\%), rather than a simple
chair/otherfurniture swap. Similar structural absorption also appears for desk and
cabinet, indicating that semantically weak or geometrically ambiguous object regions
are often overridden by dominant layout labels near boundaries.

Figure~\ref{fig:radius_presence_qual} complements the semantic mesh analysis by showing
object-level retrieval under a radius-based matching criterion. The two scenes show a
clear contrast. In scene0054, the method matches 15 of 41 GT objects at $r=0.5$ m,
corresponding to precision $= 0.4286$, recall $= 0.3659$, and $F1 = 0.3947$; many
``Missing GT'' markers and unmatched predictions remain in the cluttered layout. By
contrast, scene0233 matches 11 of 17 GT objects with only two unmatched predictions,
corresponding to precision $= 0.8462$, recall $= 0.6471$, and $F1 = 0.7333$. The
remaining failures are concentrated in cluttered multi-object regions rather than
reflecting global scene misalignment.

\FloatBarrier

\subsection{uHumans2 Dynamic Scene Ablation}
\label{sec:uhumans2_ablation}

We use uHumans2~\cite{rosinol2021kimera} to test how the proposed
perception-to-mapping components behave when the scene contains moving humans. The
ablation uses two sequence groups: Apartment H1/H2 and Office H6/H12, where H denotes
the number of moving humans. Apartment is shorter and has less accumulated drift,
whereas Office is a longer dynamic setting with more trajectory drift and more
opportunities for revisiting the same regions. All rows use the same M2H-MX perception
model, RVIO2 odometry, SuperPoint tracking, and volumetric fusion settings. The baseline
disables sparse predicted-depth factors, semantic masking, and temporal pose-warp
fusion.

Table~\ref{tab:uhumans2_component_message} compares sparse predicted-depth factors,
semantic masking, and temporal pose-warp fusion using four metrics.
ATE measures trajectory drift, Node F1 measures object-level scene-graph quality,
Chamfer measures geometric reconstruction quality, and scene-graph similarity summarizes
the agreement of the predicted graph with the reference graph. Relative changes are
reported only for ATE and Node F1, since these are the two main metrics used to interpret
the ablation. A positive ATE change means lower trajectory error, and a positive Node F1
change means better object-node graph quality.

\begin{table}[!t]
\centering
\caption{uHumans2 dynamic-scene ablation. The baseline disables sparse
predicted-depth factors, semantic masking, and temporal pose-warp fusion. Rows with
$K=0$ enable sparse predicted-depth factors and semantic masking but disable temporal
pose-warp fusion. Rows with $K \in \{1,3,5\}$ add temporal pose-warp fusion on top of
that configuration. ATE and Node F1 are the primary ablation metrics and include
relative change from the baseline of the same split. Chamfer and scene-graph similarity
are included as supporting reconstruction and graph-consistency metrics. Best values
within each split are bolded, and second-best values are underlined.}
\label{tab:uhumans2_component_message}
\scriptsize
\setlength{\tabcolsep}{3.2pt}
\resizebox{\textwidth}{!}{%
\begin{tabular}{llcccc}
\toprule
Split & Configuration &
\shortstack{ATE $\downarrow$\\(rel.)} &
\shortstack{Node F1 $\uparrow$\\(rel.)} &
Chamfer $\downarrow$ &
\shortstack{Scene graph\\similarity $\uparrow$} \\
\midrule
\multirow{7}{*}{Apartment H1/H2}
& Baseline
& 0.068 (--)
& 0.432 (--)
& 0.591
& 0.216 \\
& Depth sparse factors
& \textbf{0.051 (+25.0\%)}
& 0.466 (+7.9\%)
& \underline{0.590}
& 0.233 \\
& Semantic mask
& \textbf{0.051 (+25.0\%)}
& 0.452 (+4.6\%)
& 0.593
& 0.226 \\
& Depth sparse factors + semantic mask, $K=0$
& 0.075 (-10.3\%)
& \textbf{0.496 (+14.8\%)}
& 0.595
& \textbf{0.248} \\
& Depth sparse factors + semantic mask, $K=1$
& \underline{0.055 (+19.1\%)}
& 0.472 (+9.3\%)
& \textbf{0.587}
& 0.236 \\
& Depth sparse factors + semantic mask, $K=3$
& \textbf{0.051 (+25.0\%)}
& 0.444 (+2.8\%)
& 0.596
& 0.222 \\
& Depth sparse factors + semantic mask, $K=5$
& 0.061 (+10.3\%)
& \underline{0.483 (+11.8\%)}
& 0.593
& \underline{0.242} \\
\midrule
\multirow{7}{*}{Office H6/H12}
& Baseline
& 1.332 (--)
& 0.347 (--)
& 2.007
& 0.274 \\
& Depth sparse factors
& 1.258 (+5.6\%)
& 0.380 (+9.5\%)
& 2.070
& 0.290 \\
& Semantic mask
& 1.281 (+3.8\%)
& 0.385 (+11.0\%)
& 2.105
& 0.292 \\
& Depth sparse factors + semantic mask, $K=0$
& 1.214 (+8.9\%)
& 0.390 (+12.4\%)
& 2.084
& 0.295 \\
& Depth sparse factors + semantic mask, $K=1$
& 1.283 (+3.7\%)
& \textbf{0.452 (+30.3\%)}
& \textbf{1.506}
& \textbf{0.316} \\
& Depth sparse factors + semantic mask, $K=3$
& \textbf{1.125 (+15.5\%)}
& \underline{0.440 (+26.8\%)}
& \underline{1.696}
& \underline{0.308} \\
& Depth sparse factors + semantic mask, $K=5$
& \underline{1.195 (+10.3\%)}
& 0.383 (+10.4\%)
& 1.817
& 0.281 \\
\bottomrule
\end{tabular}
}
\end{table}

Table~\ref{tab:uhumans2_component_message} first shows that the two local-refinement cues are
useful on their own. In Apartment, sparse predicted-depth factors and semantic masking
both reduce ATE from 0.068\,m to 0.051\,m. In Office, the same cues also reduce drift,
although the gains are smaller: sparse depth lowers ATE from 1.332\,m to 1.258\,m, and
semantic masking lowers it to 1.281\,m. This pattern is consistent with their roles in
local graph refinement. Sparse predicted depth provides metric anchors at tracked keypoints, while
semantic masking removes features and depth factors from dynamic regions before they
enter local graph refinement.

Combining sparse depth and semantic masking reveals a trade-off between trajectory
and object-node accuracy. In Apartment, it raises Node F1 from 0.432 to 0.496 and improves
scene-graph similarity from 0.216 to 0.248, but the ATE increases to 0.075\,m. A likely explanation is that, in a short sequence
where the baseline trajectory is already stable, the added
constraints can improve object-level consistency without necessarily improving the local
pose estimate. In Office, where drift and dynamic objects are more severe, the same
combination reduces trajectory error and improves object-node accuracy. ATE decreases from 1.332\,m to
1.214\,m, and Node F1 increases from 0.347 to 0.390. Thus, the combined cues are most
useful when the sequence is long enough for dynamic-scene drift to accumulate.

\begin{figure}[!t]
\centering
\includegraphics[width=\textwidth]{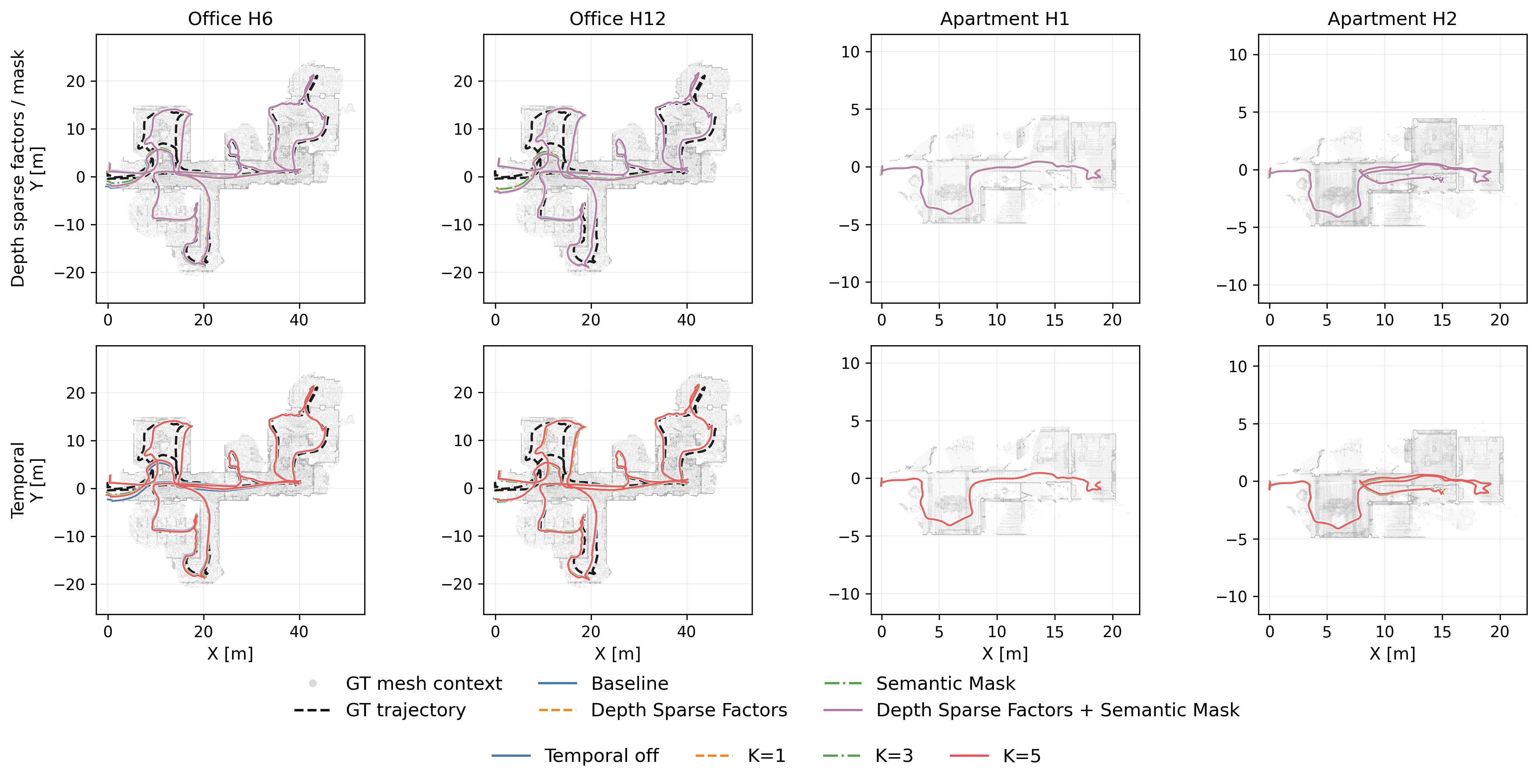}
\caption{Trajectory and mesh overview for the uHumans2 ablation. The top row summarizes
the depth-sparse-factor and semantic-masking comparison, while the bottom row summarizes
the temporal pose-warp sweep. The figure provides qualitative context for the trajectory
and mesh changes reported in Table~\ref{tab:uhumans2_component_message}.}
\label{fig:uhumans2_ablation_trajectory_mesh_overview}
\end{figure}

Figure~\ref{fig:uhumans2_ablation_trajectory_mesh_overview} compares the trajectories
and meshes for these settings. As the table shows, the setting with the lowest
trajectory error does not necessarily give the best reconstruction or object-node
accuracy.

The temporal rows show the clearest trade-off. In Office, $K=3$ gives the strongest
trajectory result, reducing ATE from 1.332\,m to 1.125\,m. This indicates that a moderate
temporal support window helps when the trajectory has accumulated drift. However, $K=1$
gives the best local reconstruction and graph indicators in Office: Chamfer decreases to
1.506, Node F1 increases to 0.452, and scene-graph similarity increases to 0.316. Short
temporal support therefore preserves local consistency well, while a wider support window
better stabilizes the trajectory. The $K=5$ row is weaker than $K=3$ for ATE and weaker
than $K=1$ for graph quality, suggesting that older warped predictions can become stale
or misaligned in dynamic scenes.

\begin{figure}[!t]
\centering
\setlength{\tabcolsep}{2pt}
\renewcommand{\arraystretch}{0.95}
\resizebox{\textwidth}{!}{%
\begin{tabular}{ccc}
\includegraphics[width=0.32\textwidth]{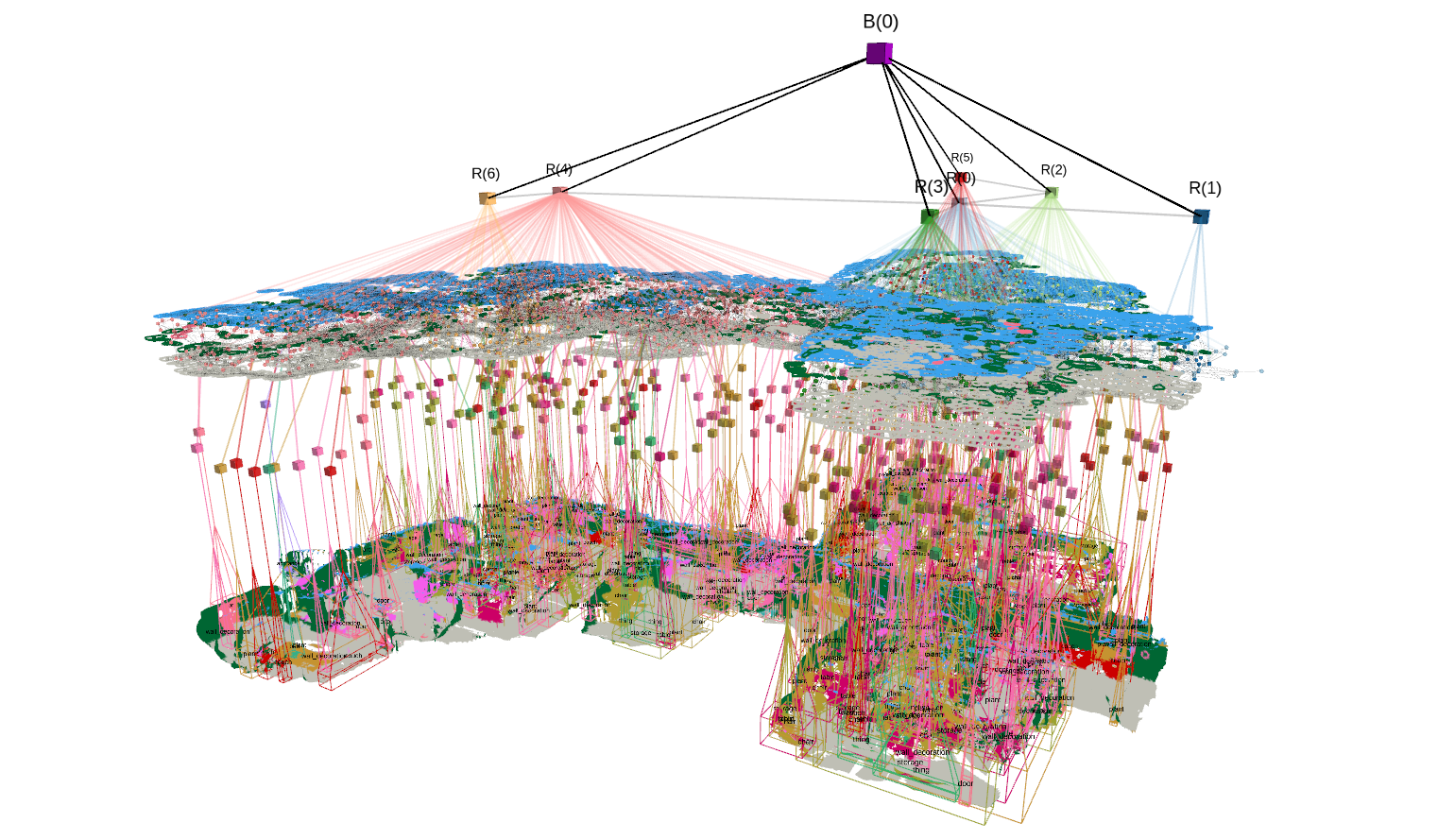} &
\includegraphics[width=0.32\textwidth]{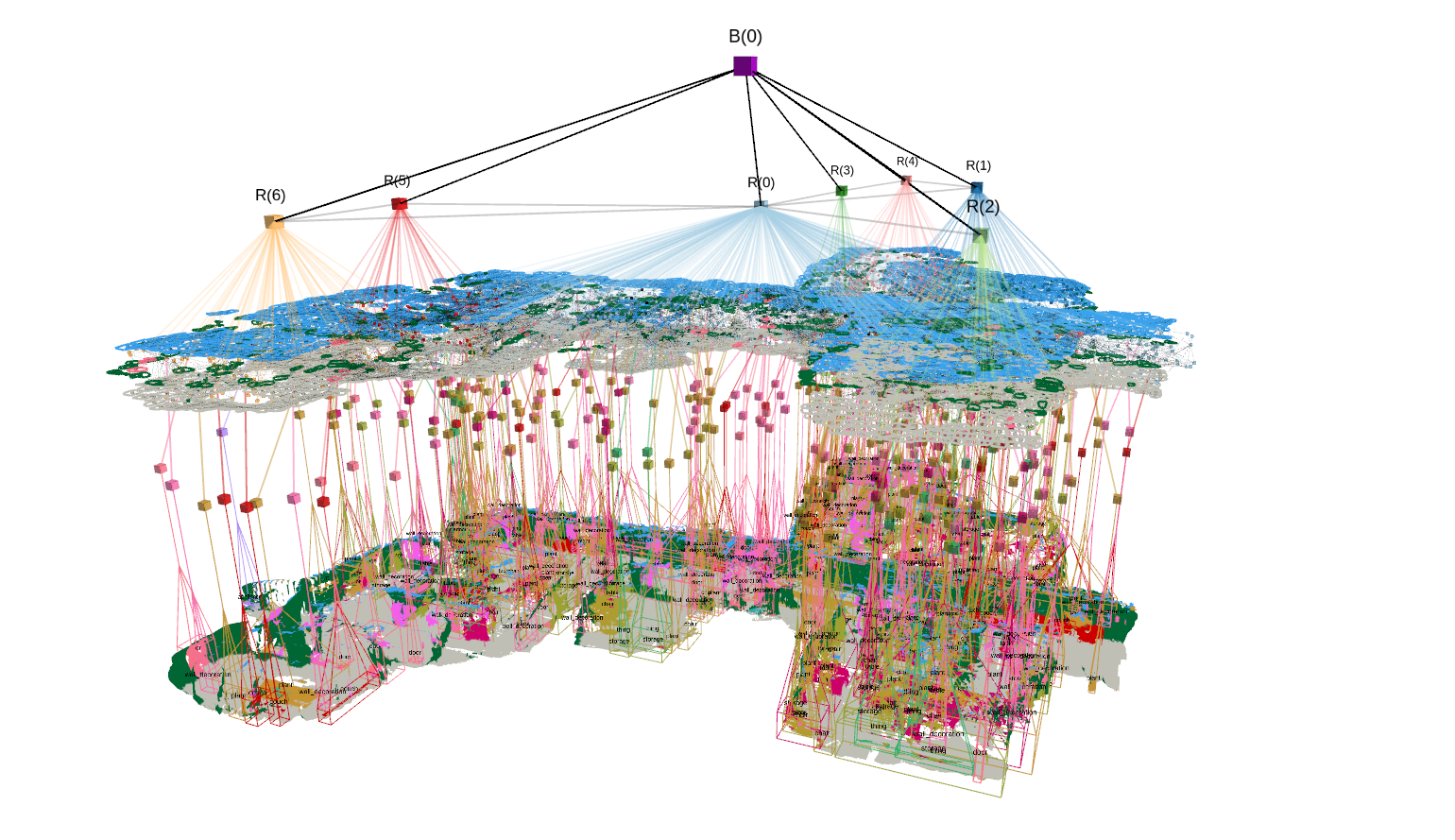} &
\includegraphics[width=0.32\textwidth]{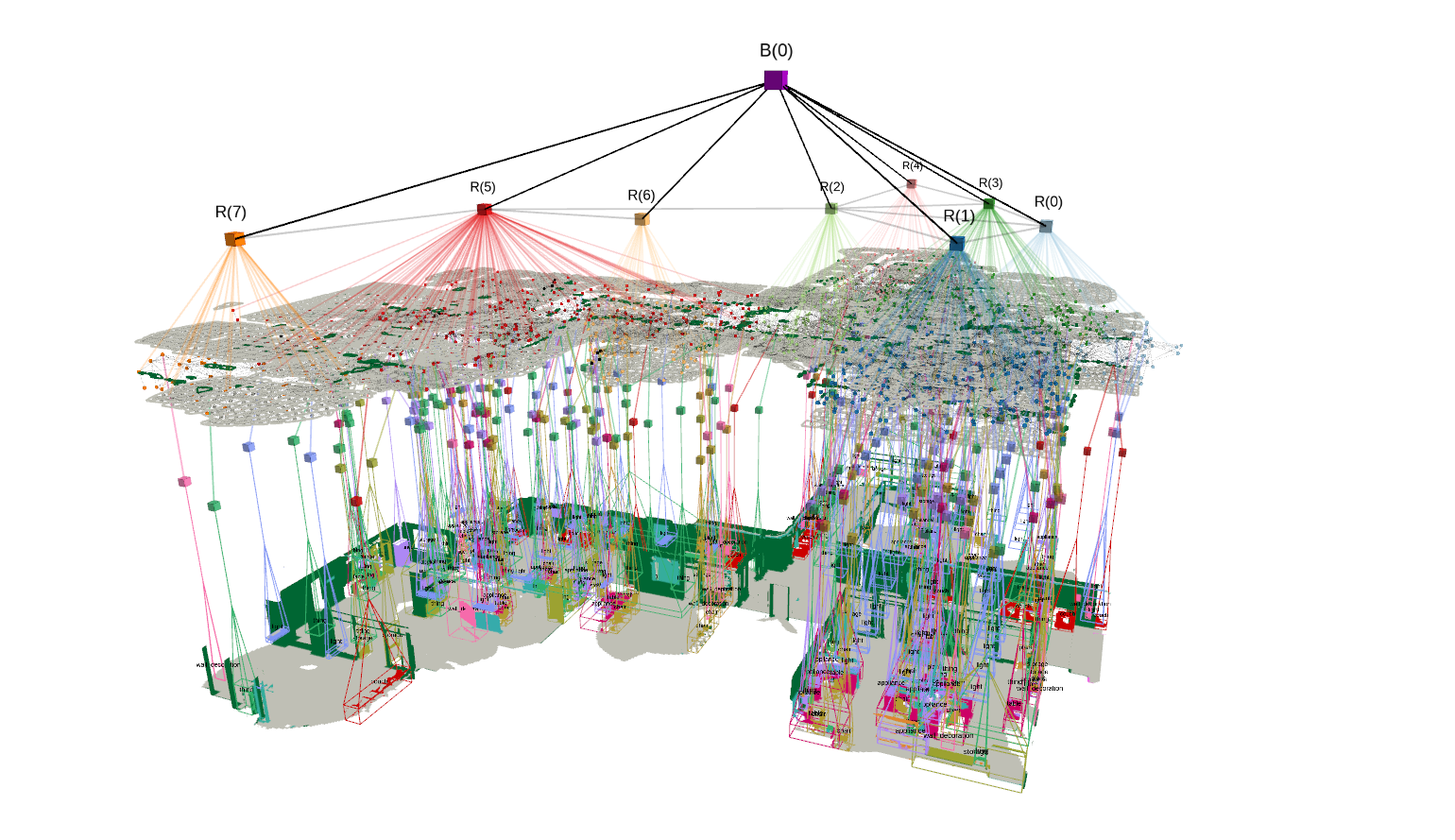} \\
{\small Baseline} & {\small Sparse depth + semantic mask} & {\small Ground truth} \\
\includegraphics[width=0.32\textwidth]{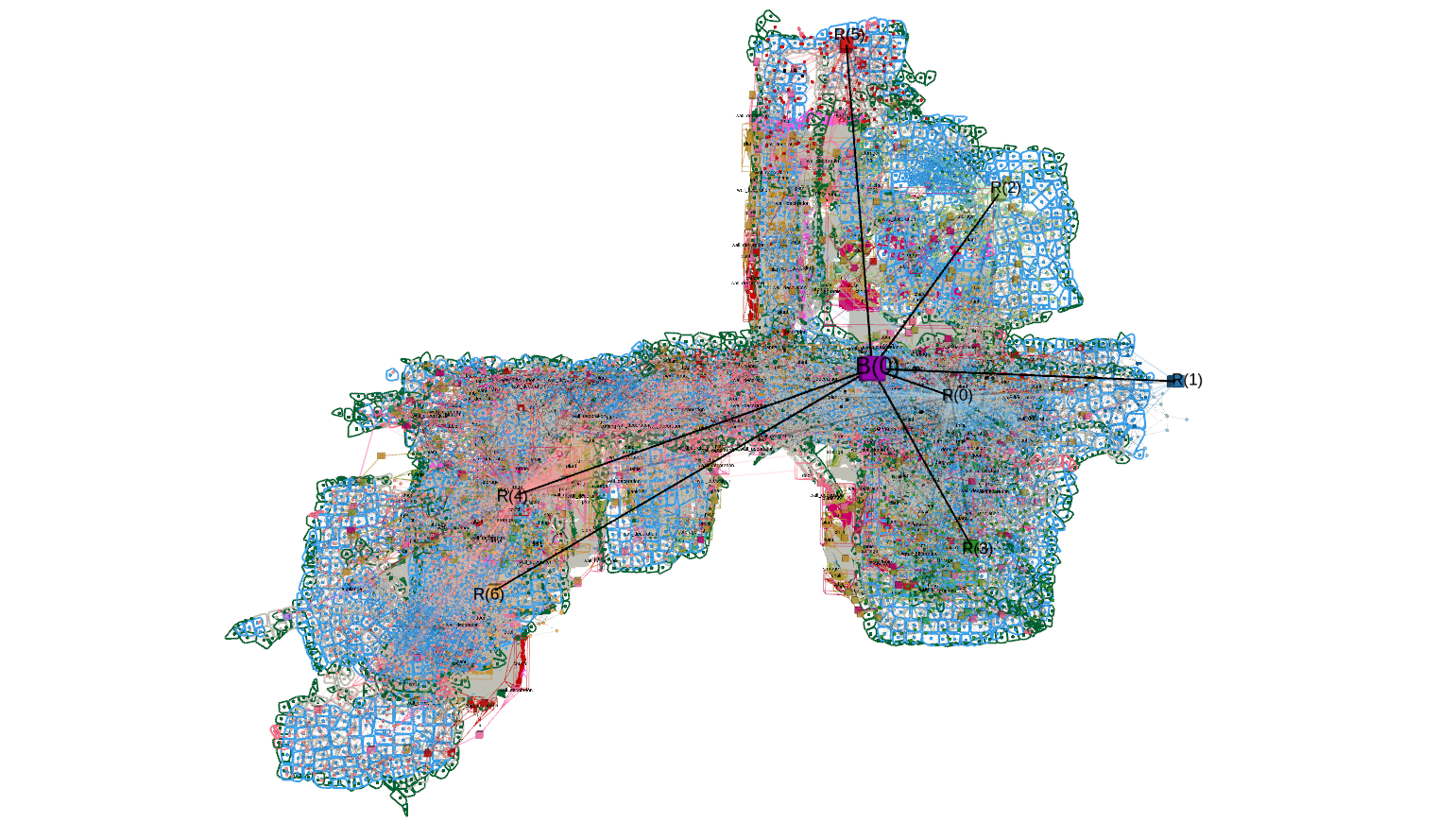} &
\includegraphics[width=0.32\textwidth]{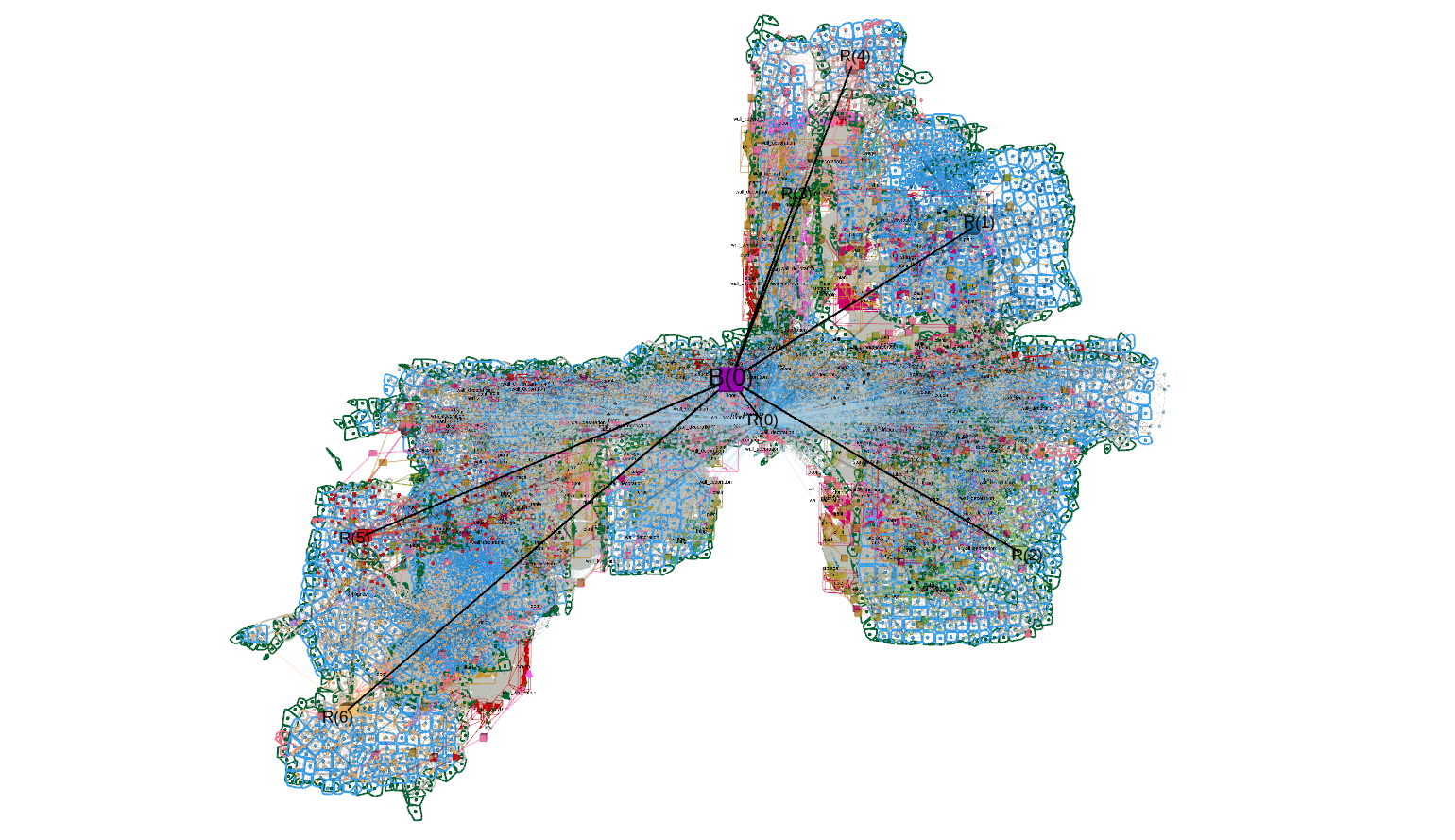} &
\includegraphics[width=0.32\textwidth]{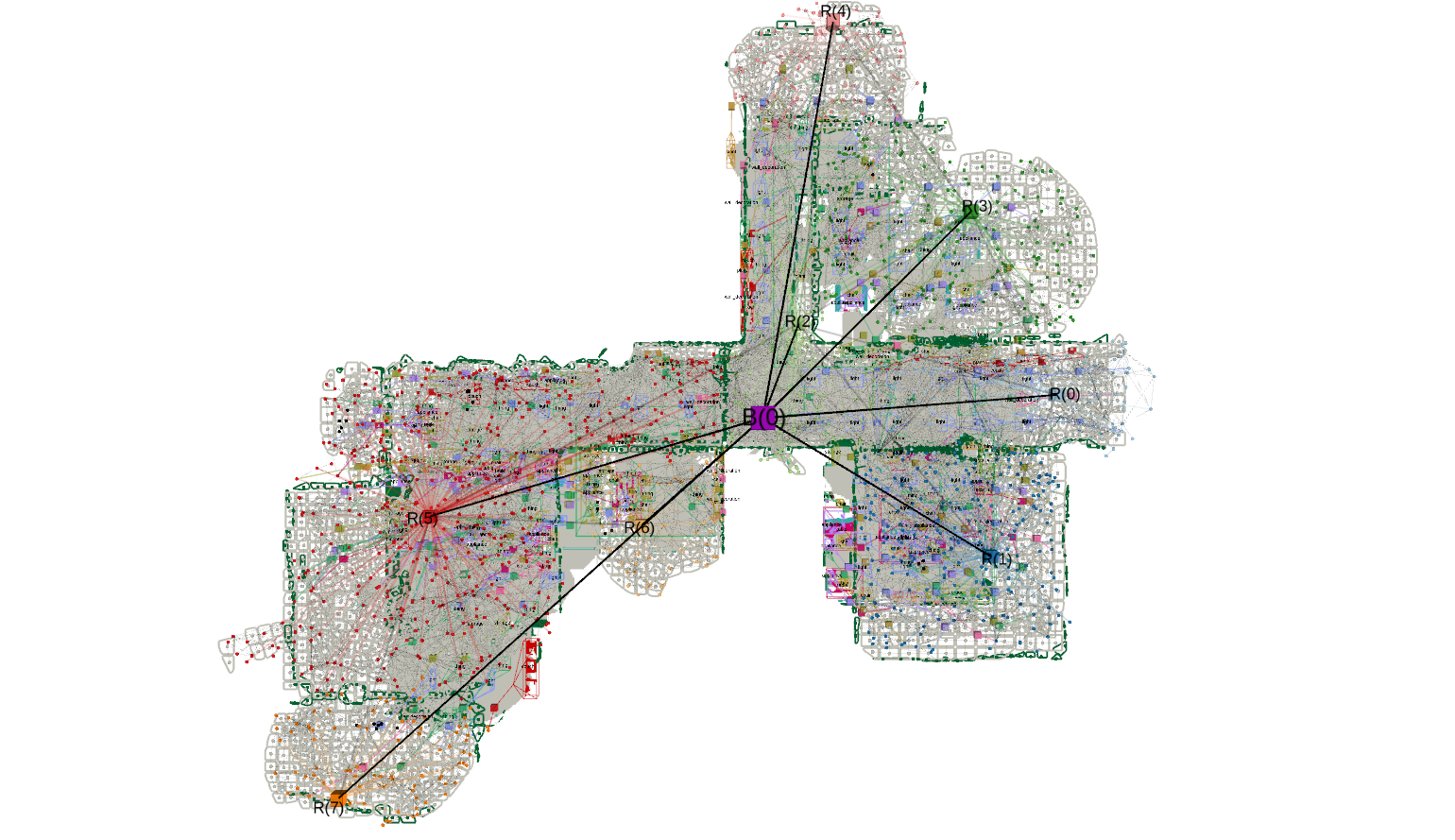} \\
{\small Baseline top view} & {\small Sparse depth + semantic mask top view} &
{\small Ground-truth top view}
\end{tabular}}
\caption{Qualitative uHumans2 Office H12 ablation example. The top row shows the
isometric reconstruction and the bottom row shows the corresponding top view for the
baseline, sparse depth + semantic masking, and the sequence-specific ground-truth
reference. The scene contains twelve moving humans and illustrates the dynamic complexity
behind the aggregate Office results in Table~\ref{tab:uhumans2_component_message}.}
\label{fig:uhumans2_office_h12_ablation}
\end{figure}

Figure~\ref{fig:uhumans2_office_h12_ablation} shows the most challenging Office case
qualitatively. The sequence contains twelve moving humans, where semantic masking and
sparse depth factors matter most: local graph refinement must retain stable static-scene constraints while avoiding
features whose motion is not explained by the camera trajectory.

\paragraph{Loop-Closure Candidate Analysis}
\label{sec:uhumans2_loop_closure}

We use loop-closure candidates to check whether the estimated trajectory remains close
enough to revisited regions for the backend to form loop closures.

\begin{figure}[!t]
\centering
\includegraphics[width=\textwidth]{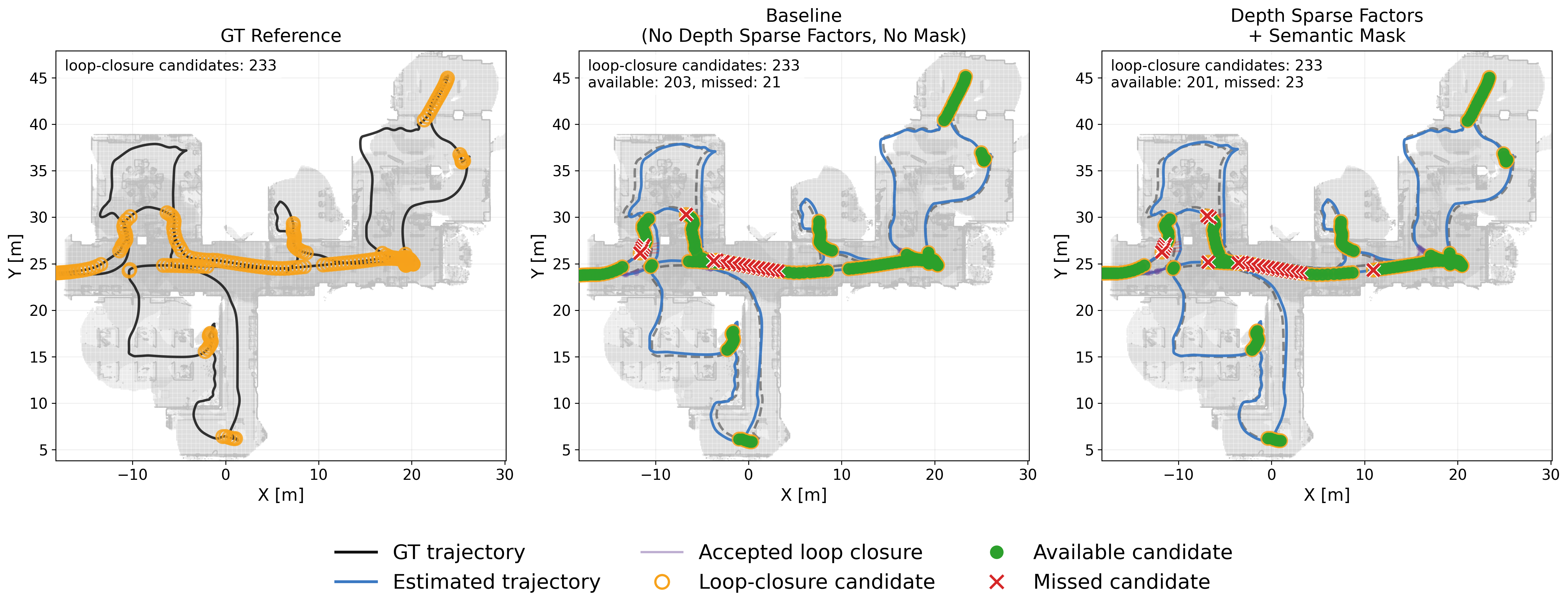}
\caption{Loop-closure candidate diagnostic on the highly dynamic uHumans2 Office H12
sequence. All panels use the same ground-truth mesh as spatial context. The two
configurations show similar revisit-candidate coverage but different trajectory drift.}
\label{fig:uhumans2_office_h12_loopclosure_diagnostic}
\end{figure}

Figure~\ref{fig:uhumans2_office_h12_loopclosure_diagnostic} shows similar revisit-candidate
coverage for the baseline and sparse-depth-plus-semantic-mask configurations in Office H12.
The latter trajectory shows less drift relative to the ground truth.

The three components improve different parts of the pipeline. Sparse predicted-depth factors constrain landmark depths in local graph refinement.
Semantic masking removes keypoints and sparse depth factors on dynamic regions before
that optimization. Temporal pose-warping aligns depth and labels across frames,
but its window length changes the balance between trajectory accuracy and graph quality.
In Office, $K=3$ gives the best ATE,
whereas $K=1$ gives the strongest Office Node F1, Chamfer, and scene-graph similarity.

\FloatBarrier

\subsection{ITC 3D Mapping Test}
We also conduct a real-world 3D mapping test on the ITC dataset
introduced in the Mono-Hydra study~\cite{udugama2023monohydra}, which
was captured in the ITC building using the RGB sensor (OmniVision OV2740) and IMU
(Bosch BMI055) data from a RealSense D435i camera. In
contrast to ScanNet and 7-Scenes, this dataset reflects a practical indoor deployment
setting with long corridor sequences, repeated structural patterns, and extended
traversals, where repetitive geometry, low-texture regions, and long-range accumulation
can increase drift and reduce reconstruction stability for monocular RGB+IMU mapping.

\begin{figure}[!t]
\centering
\begin{subfigure}[t]{0.86\linewidth}
\centering
\includegraphics[width=\linewidth]{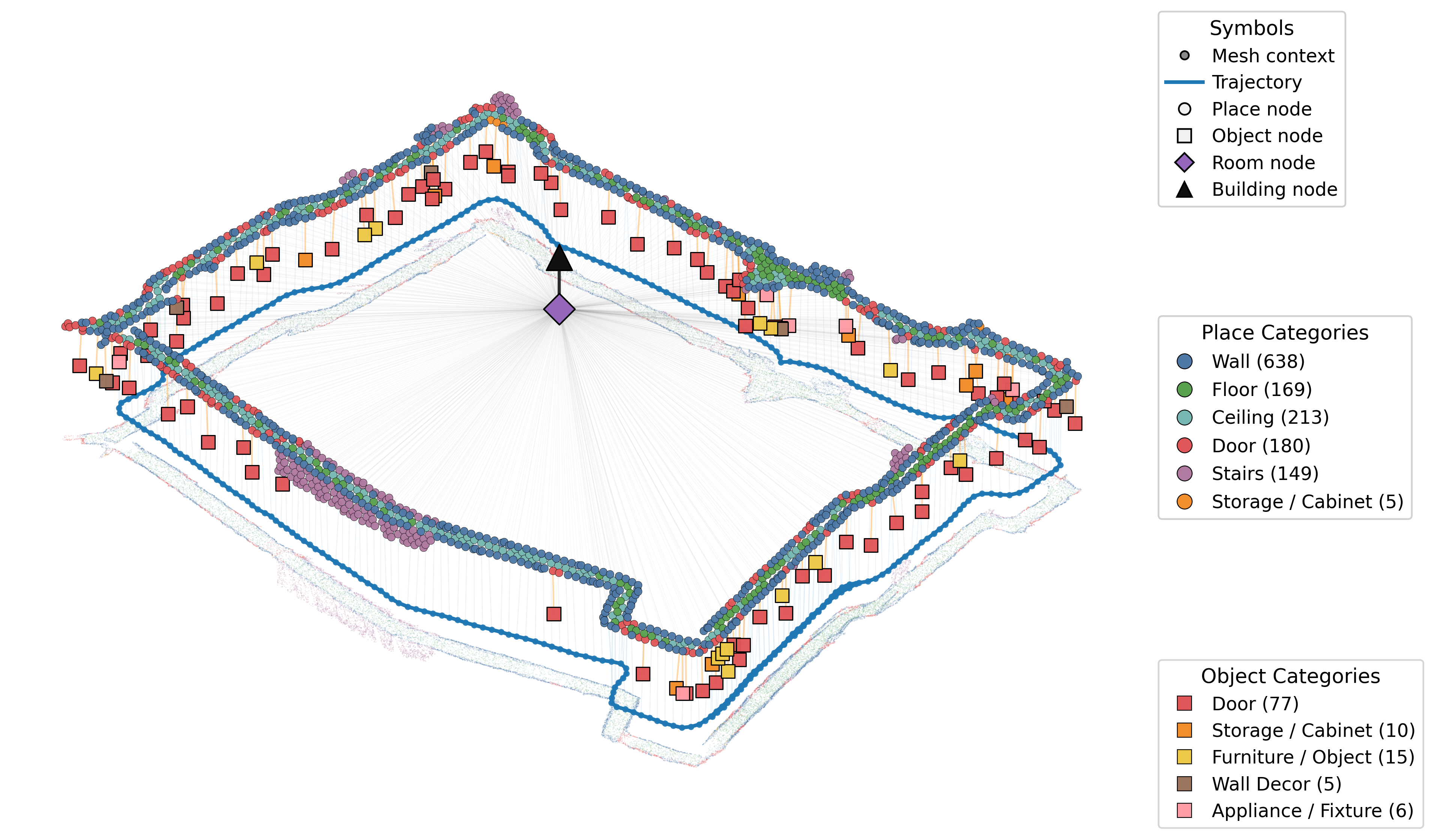}
\caption{M2H-MX-L, $640\times480$ input resolution}
\end{subfigure}

\vspace{0.5mm}

\begin{subfigure}[t]{0.86\linewidth}
\centering
\includegraphics[width=\linewidth]{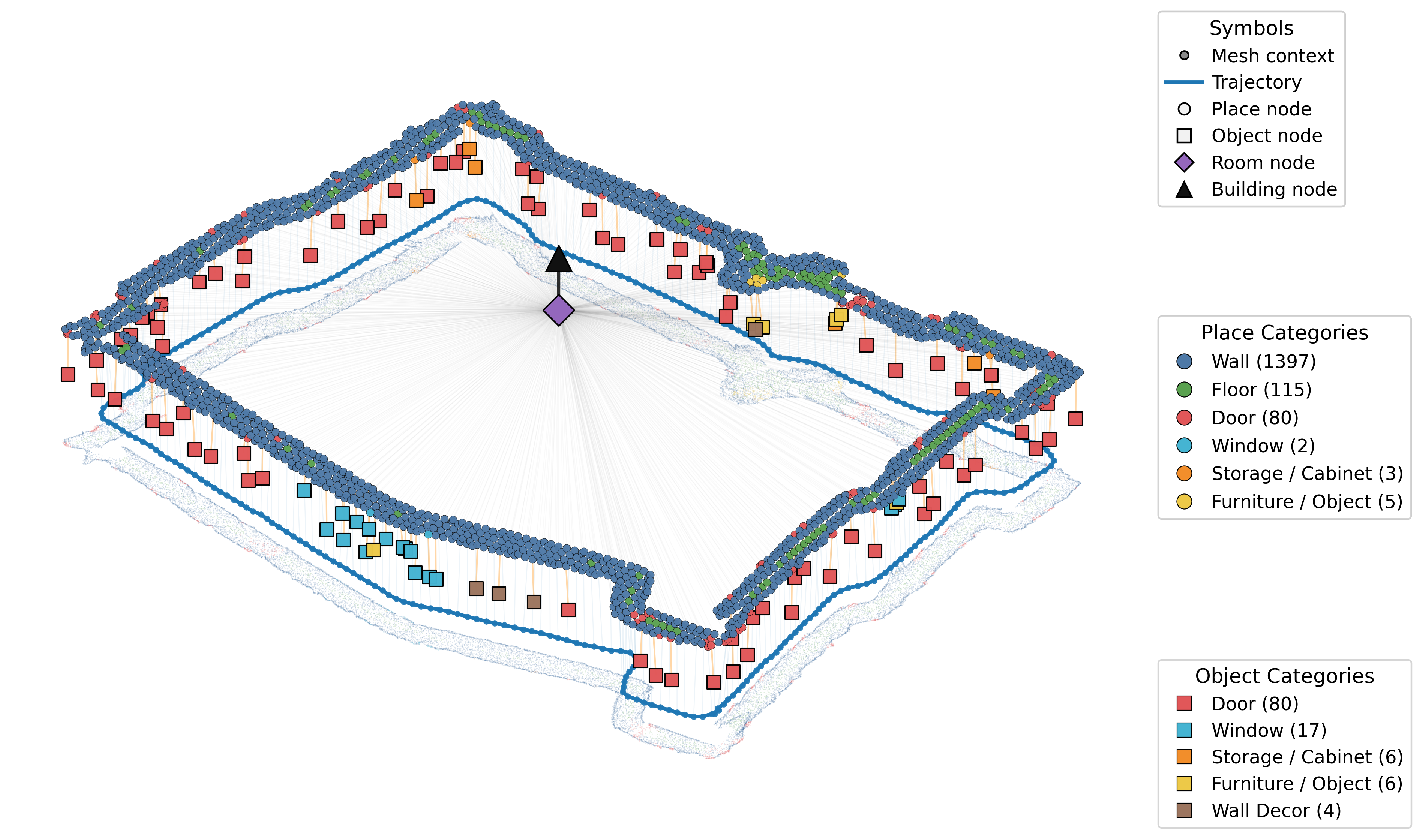}
\caption{M2H-MX-L ONNX, $256\times192$ input resolution}
\end{subfigure}
\caption{ITC 2nd Floor full-loop reconstructions of a corridor loop of roughly 200 m.
(a) Full M2H-MX-L run on an NVIDIA RTX 4080 Super 16GB; (b) ONNX deployment on a
Jetson Orin NX 16GB.}
\label{fig:itc_2ndfloor_loop}
\end{figure}

\FloatBarrier

In the Mono-Hydra study~\cite{udugama2023monohydra}, the ITC dataset was used
to assess real captured data against a LiDAR-based reference, demonstrating the
feasibility of generating metrically meaningful 3D meshes from monocular RGB+IMU input
in a real building environment.
In the present work, we use the same real-world test setting to compare mapping
accuracy with different perception models in a common downstream mapping pipeline.
Figure~\ref{fig:itc_2ndfloor_loop} compares the reconstructions obtained with the
full M2H-MX-L model and the embedded ONNX deployment. The ONNX deployment more
consistently labels the covered walking-passage walls as windows, whereas the full
model labels some regions visible through the glass as stairs, likely because of
roof-like structure behind the glass.

\begin{table}[!t]
\centering
\footnotesize
\setlength{\tabcolsep}{6pt}
\renewcommand{\arraystretch}{1.05}
\caption{3D mapping results on the ITC real-world dataset introduced in the original Mono-Hydra
study~\cite{udugama2023monohydra}, captured in the ITC building using a RealSense
camera. ME and SD denote the mean and standard deviation of the mapping error in
meters. Lower values are better.}
\label{tab:itc_mapping}
\begin{tabular}{lcccc}
\toprule
& \multicolumn{2}{c}{2nd Floor} & \multicolumn{2}{c}{3rd Floor} \\
\cmidrule(lr){2-3}\cmidrule(lr){4-5}
Model & ME (m)$\downarrow$ & SD (m)$\downarrow$ & ME (m)$\downarrow$ & SD (m)$\downarrow$ \\
\midrule
Mono-Hydra~\cite{udugama2023monohydra} & 0.19 & 0.18 & 0.21 & 0.16 \\
MTMamba++~\cite{lin2025mtmamba++} & 0.21 & 0.22 & 0.18 & 0.19 \\
M2H~\cite{udugama2025m2h} & 0.11 & 0.14 & 0.10 & 0.13 \\
M2H-MX-L (Jetson ONNX, $256\times192$) & 0.22 & 0.16 & 0.19 & 0.18 \\
M2H-MX-B & 0.10 & 0.11 & 0.09 & 0.10 \\
M2H-MX-L & 0.08 & 0.08 & 0.07 & 0.09 \\
\bottomrule
\end{tabular}
\end{table}

Table~\ref{tab:itc_mapping} summarizes the 3D mapping performance on the ITC
real-world dataset. The evaluation is based on mean error and standard deviation of the
mapping error in meters, with lower values indicating better reconstruction accuracy
and stability. The full-resolution M2H-MX-B and M2H-MX-L variants outperform the
earlier baselines, and M2H-MX-L achieves the best overall results on both floors.
Because Mono-Hydra uses separate single-task depth and semantic networks, this table
also compares separate single-task networks with joint multi-task prediction in the same
mapping setting: M2H-MX-L reduces the mean mapping error from 0.19\,m to 0.08\,m on the
2nd floor and from 0.21\,m to 0.07\,m on the 3rd floor.
These results indicate that M2H-MX perception also improves real-world
mapping performance in long-corridor indoor environments. The main ITC mapping tests were conducted on an NVIDIA RTX 4080 Super 16GB,
where the two-head M2H-MX-L deployment uses the depth and semantic outputs
required by the mapping system and sustains $25$--$30$~Hz in the asynchronous
perception-to-mapping loop at $640\times480$ input resolution. To assess embedded
deployment, we also ran an ONNX-exported M2H-MX-L model on a Jetson Orin NX 16GB with
TensorRT FP16 and CUDA Graphs at $256\times192$ input resolution. This configuration
reaches about 25 FPS for the deployed perception model. Its reconstruction errors are reported in the Jetson ONNX row of
Table~\ref{tab:itc_mapping}. The embedded ONNX configuration, at its lower $256\times192$ input
resolution, is less accurate than the full-resolution M2H-MX-B/L evaluations on the ITC
mapping benchmark.

\begin{table}[!p]
\centering
\caption{Component timings and memory utilization of Mono-Hydra++ on the ITC dataset, by
platform, execution backend, and network input resolution. Processing times are mean $\pm$ SD in ms per image, prediction, keyframe, or mapping update, as indicated. Memory utilization is the mean share of installed capacity: 32 GiB system RAM and 16 GiB GPU memory on the workstation, and 16 GiB memory shared by CPU and GPU on the Jetson.}
\label{tab:runtime}
\setlength{\tabcolsep}{3pt}
{\footnotesize
\begin{tabular}{@{}>{\raggedright\arraybackslash}p{0.37\textwidth}rrr@{}}
\toprule
Component / invocation & \shortstack{RTX 4080 Super\\CUDA\\$640\times480$} & \shortstack{RTX 4080 Super\\CUDA\\$256\times192$} & \shortstack{Jetson Orin NX\\TensorRT FP16\\$256\times192$} \\
\midrule
\multicolumn{4}{@{}l}{\textit{Processing time (ms)}} \\
Joint depth + semantic prediction / image & $36.03\pm2.24$ & $15.73\pm0.95$ & $34.29\pm0.63$ \\
SuperPoint extraction / image & $10.40\pm1.63$ & $9.99\pm1.23$ & $33.88\pm5.89$ \\
Complete VIO processing / image & $13.97\pm1.96$ & $13.55\pm1.60$ & $34.69\pm5.80$ \\
Local refinement / keyframe & $18.70\pm2.27$ & $19.05\pm2.23$ & $93.20\pm9.45$ \\
Temporal fusion / prediction & $40.14\pm3.07$ & $20.06\pm1.60$ & $37.45\pm2.66$ \\
Loop-closure processing / keyframe & $14.59\pm8.63$ & $16.84\pm9.26$ & $43.98\pm23.44$ \\
Scene-graph frontend / update & $313.92\pm35.04$ & $371.89\pm19.06$ & $545.45\pm36.02$ \\
Backend: sum of sequential timed regions / update & $481.52\pm93.91$ & $407.88\pm101.55$ & $1068.47\pm230.14$ \\
\midrule
\multicolumn{4}{@{}l}{\textit{Resource use}} \\
Memory utilization, RAM / GPU (\%)\textsuperscript{1} & 52.7 / 28.5 & 41.5 / 14.6 & 46.9 (shared) \\
\bottomrule
\end{tabular}
}
\smallskip
\begin{minipage}{\textwidth}
\footnotesize\textit{Notes:} One image denotes one input frame; one prediction
denotes one depth--semantic output pair. Nested timings, such as SuperPoint extraction within VIO,
overlap.
Asynchronous stage timings are not additive end-to-end latency.
\textsuperscript{1}System RAM includes memory used outside the pipeline.
\end{minipage}
\end{table}

A separate ITC test compares the computational performance of the complete
Mono-Hydra++ stack on the RTX 4080 Super workstation and Jetson Orin NX 16GB
using similar RGB and IMU inputs. On Jetson, this test uses the ONNX-exported
M2H-MX-L model at $256\times192$ network input, as in the reconstruction test
reported in Table~\ref{tab:itc_mapping}. The RTX 4080 Super workstation also uses
M2H-MX-L at $256\times192$ to compare timings at the same input resolution,
with CUDA on the workstation and TensorRT FP16 on Jetson.
Table~\ref{tab:runtime} also reports the RTX 4080 Super at the $640\times480$ input resolution used for the main ITC mapping runs.
Table~\ref{tab:runtime} reports component processing times and
resource use for this comparison. Network prediction, SuperPoint extraction, VIO, local refinement,
temporal fusion, and loop-closure processing are timed alongside
scene-graph updates. The backend entry sums its sequential timed regions per
update. VIO tracking and scene-graph updates run asynchronously, with the graph
updated less frequently; these invocation times are not additive end-to-end latency.

\FloatBarrier
\section{Discussion and Limitations}

The experiments show that monocular RGB+IMU sensing can support real-time
metric-semantic mapping and hierarchical 3D scene-graph construction. RGB-D baselines use measured depth
and are included as sensor-rich references, whereas Mono-Hydra++ uses RGB and IMU
without RGB-D or LiDAR input. On the selected ScanNet sequences, the RGB+IMU pipeline
achieves trajectory accuracy comparable to the strongest depth-based system, with a
1.6\% lower average ATE. Temporal fusion also contributes to these results: in the Office
ablation, adding a short temporal window improves reconstruction and object-node
accuracy over the same depth-and-mask configuration without temporal fusion
(Table~\ref{tab:uhumans2_component_message}). The RGB-D comparisons therefore assess
the complete systems, including their different sensing inputs and temporal processing.

The main remaining limitation is object-level semantic preservation. The semantic mesh
and uHumans2 analyses show that large structural classes and large objects are recovered
more reliably than the lower-IoU object categories in the classwise evaluation. These
failures can remove the semantic evidence needed to form object nodes and may affect
object-room assignments or structural relations. The reported object and graph metrics
do not separately establish room-assignment accuracy or object persistence over time.
Improving graph quality therefore requires preserving object evidence during semantic
fusion as well as controlling pose error.

Possible mitigations are to retain uncertain object labels until additional views
resolve them, downweight depth samples where predicted boundaries disagree across views,
and check whether an object is visible before updating its node. These are directions
for future work, to be evaluated using object retention, false-node, and room-relation
tests. The present depth-only and masking-only ablations separate the effects of sparse
depth and semantic masking; the comparison with separate single-task networks is
limited to the ITC mapping test.

Temporal pose-warp fusion also depends on the quality of the VIO trajectory. Short
windows improve reconstruction and object-node accuracy in Office, but longer
windows can introduce outdated semantic evidence when pose estimates drift or dynamic
objects occupy the scene. In practice, the window length should follow the intended
use: a short window when map and scene-graph consistency matter most, and a moderate
window when trajectory drift dominates. Adapting the window to estimated pose quality
is a possible extension that remains to be evaluated. Semantic masking also depends
on the predicted labels and the configured dynamic classes; moving objects outside
those classes can still contribute geometric constraints. Calibration errors and domain shifts are additional
risks for learned-depth alignment, and their effects are not isolated by the present
evaluation.

The desktop RTX 4080 Super deployment sustains a 25--30~Hz
perception-to-mapping loop at the full $640{\times}480$ input resolution, whereas the
Jetson Orin NX deployment achieves about 25 FPS for perception at the reduced
$256{\times}192$ input resolution. Table~\ref{tab:runtime} reports ITC component timings and resource use;
these invocation-level measurements are distinct from the deployment throughput
figures.

\FloatBarrier
\section{Conclusion}

We presented Mono-Hydra++, a monocular RGB+IMU pipeline that combines M2H-MX
perception, RVIO2 odometry with SuperPoint tracking, local graph refinement, and pose-aware
temporal alignment. Sparse predicted-depth factors and semantic filtering refine the
local graph, whose pose constraints are supplied to the Mono-Hydra/Hydra backend.
Temporal alignment stabilizes dense depth and semantic observations before volumetric
fusion and hierarchical scene-graph construction.

The dynamic-scene ablations show that the setting with the lowest trajectory error
can differ from the setting with the best reconstruction and object-node accuracy.
Temporal-window selection should therefore consider the quality of the resulting map
and graph as well as the trajectory.

Future work will extend the framework in three directions: open-set and
uncertainty-aware semantics to handle previously unseen categories while preserving
small-object evidence under noisy dense predictions, graph-level correction and
language grounding to recover missing object nodes and connect textual queries to
objects, rooms, and relations in the scene graph, and downstream navigation on
resource-constrained platforms where the scene graph supports planning and decision
making under limited onboard compute.

\clearpage
\section*{Code and data availability}

The source code, configuration files, and instructions for running Mono-Hydra++
are available at \url{https://github.com/BavanthaU/mono-hydra-pp.git}.
All public benchmark and deployment datasets used in this work are available
from their respective project pages.

\section*{Declaration of generative AI and AI-assisted technologies in the manuscript preparation process}

During the preparation of this work, the authors used ChatGPT (OpenAI, GPT-6 Astra)
to convert author-prepared TikZ diagrams into the illustrations shown in
Figs.~\ref{fig:pipeline_overview}--\ref{fig:temporal_alignment}.
The authors reviewed and edited the generated illustrations and take full
responsibility for the content of the published article.

\clearpage
\bibliographystyle{elsarticle-num-names}
\bibliography{main}

\end{document}